\pdfoutput=1

\documentclass[10pt, journal,twoside]{IEEEtran}

\ifCLASSINFOpdf
\else
\fi

\usepackage[nosort]{cite}
\usepackage{amsmath}
\usepackage{amsfonts}
\usepackage{algorithm}
\usepackage{algorithmic}
\usepackage{threeparttable}
\usepackage{graphicx}
\usepackage{subfigure}
\usepackage{float}
\usepackage{booktabs}
\usepackage{makecell}
\usepackage{multirow}
\usepackage{pifont}
\usepackage{amssymb}
\usepackage{mathrsfs}
\usepackage{tablefootnote}
\usepackage{soul}
\usepackage{ulem}
\usepackage{float}

\usepackage{comment}
\usepackage{url}
\usepackage[table,xcdraw,dvipsnames]{xcolor}  
\usepackage{color}
\definecolor{citecolor}{RGB}{66,168,235}
\definecolor{linkcolor}{RGB}{255,0,0}
\usepackage{hyperref}
\hypersetup{colorlinks=true,citecolor=citecolor,linkcolor=linkcolor,urlcolor=Thistle}
\usepackage[utf8]{inputenc}
\usepackage{pgfplots}
\usepackage{pgfplotstable}
\pgfplotsset{compat=newest}

\usepackage{colortbl}

\begin{document}

\title{
A Billion-scale Foundation Model\\for Remote Sensing Images
}

\author{Keumgang Cha,
        Junghoon Seo,
        Taekyung Lee

\thanks{Manuscript received. \textit{(Corresponding author: Keumgang Cha)}}
\thanks{The authors are with SI Analytics, Daejeon 34051, South Korea (e-mail: chagmgang@si-analytics.ai; jhseo@si-analytics.ai; greatlight@si-analytics.ai).}
}%

\markboth{IEEE Journal of Selected Topics in Applied Earth Observations and Remote Sensing}%
{Shell \MakeLowercase{\textit{et al.}}: A Sample Article Using IEEEtran.cls for IEEE Journals}

\maketitle

\begin{abstract}

  As the potential of foundation models in visual tasks has garnered significant attention, pretraining these models before downstream tasks has become a crucial step. The three key factors in pretraining foundation models are the pretraining method, the size of the pretraining dataset, and the number of model parameters. Recently, research in the remote sensing field has focused primarily on the pretraining method and the size of the dataset, with limited emphasis on the number of model parameters. This paper addresses this gap by examining the effect of increasing the number of model parameters on the performance of foundation models in downstream tasks such as rotated object detection and semantic segmentation. We pretrained foundation models with varying numbers of parameters, including 86M, 605.26M, 1.3B, and 2.4B, to determine whether performance in downstream tasks improved with an increase in parameters. To the best of our knowledge, this is the first billion-scale foundation model in the remote sensing field. Furthermore, we propose an effective method for scaling up and fine-tuning a vision transformer in the remote sensing field. To evaluate general performance in downstream tasks, we employed the DOTA v2.0 and DIOR-R benchmark datasets for rotated object detection, and the Potsdam and LoveDA datasets for semantic segmentation. Experimental results demonstrated that, across all benchmark datasets and downstream tasks, the performance of the foundation models and data efficiency improved as the number of parameters increased. Moreover, our models achieve the state-of-the-art performance on several datasets including DIOR-R, Postdam, and LoveDA.

\end{abstract}


\IEEEpeerreviewmaketitle

\section{introduction}\label{sec:intro}

\IEEEPARstart{R}EMOTE sensing has become an essential field for various applications, and researchers have sought to address image analysis problems using deep learning methods \cite{zhu2017deep,irwansyah2023deep,ma2019deep,zheng2020review,yuan2021review,su2020hq,khelifi2020deep}.
To address remote sensing problems with deep learning applications, it is crucial to formulate the problem well in a deep learning context and obtain high-quality data.
In order to create high-quality data, many remote sensing experts explore areas they want to detect or classify and collect high-quality imagery. After defining the properties of the target to detect, objects in the collected imagery are labeled, and the model is trained. However, unlike in fields dealing with natural images, this process is challenging in remote sensing. Firstly, some objects of interest appeared too rarely to be detected, and the number of times they are exposed and remotely detected is minimal. As a result, in order to collect imagery, it is necessary to know in advance the time and place where the objects appear, which is nearly impossible. Secondly, many remote sensing experts are needed to obtain imagery and label data in a short time, making the time required to create a deep learning model comparatively longer than in other fields.

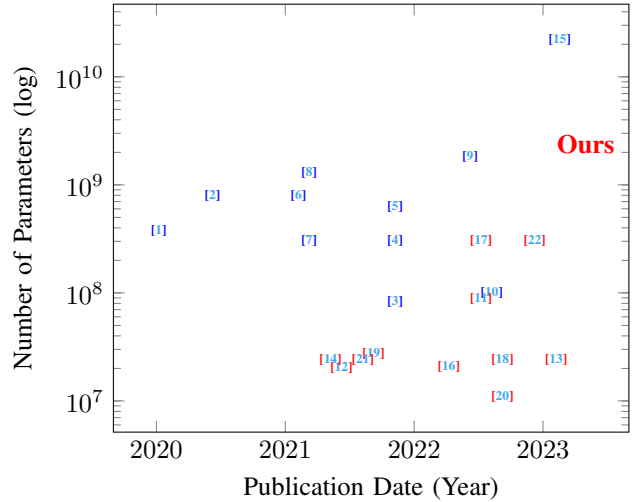
\begin{figure}
    \centering
    \begin{tikzpicture}
        \begin{axis}[ymode=log,xlabel=metri,ylabel=metri,xlabel=Publication Date (Year),ylabel=Number of Parameters (log),every axis label/.append style={font=\tiny},
            x tick label style={
                /pgf/number format/set thousands separator={},
                font=\tiny
            },
            y tick label style={font=\tiny},
              xlabel style={font=\tiny},ylabel style={font=\tiny},legend style={font=\tiny},]
            \addplot+[
                    visualization depends on={value \thisrow{nodes}\as\myvalue},
                    scatter/classes={
                    a={mark=text,text mark={\bf\tiny\myvalue},blue},
                    b={mark=text,text mark={\bf\tiny\myvalue},red},
                    c={mark=text,text mark={\bf\tiny\myvalue},red}
                    },
                    scatter,draw=none,
                    scatter src=explicit symbolic]
             table[x=x,y=y,meta=label]
                {data.txt};
        \end{axis}
    \end{tikzpicture}
    \caption{The given figure shows the variation in the size of foundation models over the years, with blue and red representing the number of parameters in computer vision and remote sensing models, respectively. While the billion-scale foundation model is already being studied in computer vision, it has not yet been developed in remote sensing. More detailed information can be found \autoref{tab:params_vs_model}. Models with fewer than 1 billion parameters are omitted.}
    \label{fig:param_year}
\end{figure}

\begin{table}[h]
     \centering
     \caption{the detailed information of the models indicated \autoref{fig:param_year}. With the exception of the model introduced in this paper, the remote sensing models are handled with the million-scale, while computer vision field has already surpassed the billion-scale in recent models.}
     \begin{tabular}{c|c|l|l}
        \hline
         Field & Referred as & Model Base & \# of Parameters \\ \hline
         \multirow{11}{*}{CV} & BYOL \cite{grill2020bootstrap} & ResNet200 2x & 375 Million \\ \cline{2-4}
          & SimCLR v2 \cite{chen2020big} & ResNet152 3x w sk & 795 Million \\ \cline{2-4}
          & DINO \cite{caron2021emerging} & ViT Base & 84 Million \\ \cline{2-4}
          & iBOT \cite{zhou2021ibot} & ViT Large & 307 Million  \\ \cline{2-4}
          & MAE \cite{he2022masked} & ViT Huge & 632 Million \\ \cline{2-4}
          & ALIGN \cite{jia2021scaling} & EfficientNet-L2 & 800 Million \\ \cline{2-4}
          & CLIP \cite{radford2021learning} & ViT Large & 307 Million \\ \cline{2-4}
          & SEER \cite{goyal2021self} & RegNety-256gf & 1.3 Billion \\ \cline{2-4}
          & Scaling ViT \cite{zhai2022scaling} & ViT Giant/14 & 1.8 Billion \\ \cline{2-4}
          & 22B ViT \cite{dehghani2023scaling} & ViT 22B & 22 Billion \\ \hline
         \multirow{12}{*}{RS} & Advanced \cite{wang2022advancing} & ViTAE-Base & 89 Million \\ \cline{2-4}
          & RingMo \cite{sun2022ringmo} & Swin Base & 88 Million \\ \cline{2-4}
          & Geograph \cite{ayush2021geography} & ResNet50 & 24 Million \\ \cline{2-4}
          & SSL in Domain \cite{ghanbarzade2023self} & ResNet50 & 24 Million\\ \cline{2-4}
          & SeCo \cite{manas2021seasonal} & ResNet50 & 24 Million \\ \cline{2-4}
          & JointSAREO \cite{wang2022self} & ViT Small & 21 Million \\ \cline{2-4}
          & SatMAE \cite{cong2022satmae} & ViT Large & 307 Million\\ \cline{2-4}
          & MM Contrastive \cite{jain2022multimodal} & ResNet50 &  24 Million \\ \cline{2-4}
          & SAR-EO \cite{cha2021contrastive} & ResNext50 & 25 Million \\ \cline{2-4}
          & Image-text \cite{mikriukov2022deep} & ResNet18 & 11 Million \\ \cline{2-4}
          & Image-audio \cite{heidler2023self} & ResNet50 &  24 Million \\ \cline{2-4}
          & Scale-MAE \cite{reed2022scale} & ViT Large &  307 Million \\ \cline{2-4}
          & EarthPT \cite{smith2023earthpt} & ViT Huge & 700 Million \\ \cline{2-4}
          & Cross-Scale MAE \cite{tang2023cross} & ViT Base & 86 Million \\ \cline{2-4}
          & Prithvi \cite{jakubik2023foundation} & ViT Base & 100 Million \\ \cline{2-4}
          & \textbf{Ours} & ViT G12$\times$4 & 2.4 Billion \\ \hline
     \end{tabular}
     \label{tab:params_vs_model}
\end{table}

{Due to the limited amount of accessible data with label compared to other fields, many studies have used fine-tuning methods with backbones trained on large amounts of data from natural images such as ImageNet, known for its ability to extract good represenstations \cite{isprs-archives-XLIII-B3-2022-1399-2022,marmanis2015deep, pires2019convolutional,liu2018classifying,huang2019and,gadiraju2023application}. However, since models pretrained with natural images such as ImageNet have a domain gap with remote sensing imagery, a large amount of high-quality labeled data is still required to make sufficient performance. Although models with enough labeled data demonstrate performance meeting requirements, they cannot approach multiple problems, a chronic issue of supervised learning. Consequently, addressing just one problem requires significant expenditure of economic and time resources.}

To tackle this issue, foundation models have been proposed in the computer vision field. Foundation models can be obtained by training a model with many parameters using unlabeled images or multiple modalities that describe a single situation. This is also called pretraining, and through this process, the foundation model can extract representations without {label}\cite{jaiswal2020survey, chen2023vlp, gan2022vision}. When fine-tuning with the foundation model as an initial point, remarkable results are achieved. In simple tasks like object classification, fine-tuning the foundation model with only a few labeled data can yield higher performance than training with all labeled data and a basic model\cite{grill2020bootstrap, chen2020simple, stojnic2021self, li2022global}. Because the property of the pretraining data is in-domain, it can result in higher performance in various tasks than pretraining on natural images and fine-tuning\cite{wang2022self, jain2022multimodal, chen2022semantic, wang2022selfreview}. Moreover, as the number of parameters constituting the model increases, performance improves gradually during fine-tuning\cite{goyal2021self,zhai2022scaling,dehghani2023scaling}. In other words, when pretraining the foundation model, the most crucial factors are the pretraining method, the amount of pretraining data, and the number of parameters constituting the model.

Following these achievements, numerous studies have been conducted on how to pretrain and create remote sensing domain-oriented foundation models. Typically, the pretraining methods for creating a foundation model in remote sensing include the following. The first is pretraining with images from a single sensor using fundamental methods such as contrastive learning \cite{jain2022multimodal}, masked image modeling \cite{wang2022advancing, sun2022ringmo, reed2022scale}, and self-distillation\cite{wang2022self}. The second method involves using multi-modal information represented by audio, text, and imagery in computer vision, but geo-location \cite{ayush2021geography}, audio \cite{heidler2023self}, and multi-spectral imagery \cite{cong2022satmae, stojnic2021self} in remote sensing. Pretraining with multi-modal information demonstrates better performance than using a single modality.

Nonetheless, most studies related to pretraining in remote sensing have focused primarily on the size of the pretraining dataset or the learning method, making it difficult to find research results on the size of the foundation model, such as \cite{sun2022ringmo}. In the existing remote sensing field, foundation model research typically deals with ResNet50 (25.6M), ResNet101 (44.5M), and vision transformer Base (86M) models\cite{he2016deep, dosovitskiy2020image}. As shown in \autoref{fig:param_year}, the number of parameters in remote sensing models is considered to be relatively small compared to those in computer vision.

{This paper addresses the extent of the number of model parameters, which is the last remaining element for building foundation models in remote sensing. For the first time, we deal with the billion-scale model in remote sensing, and prove the effect of increasing the size of the model from million-scale to billion-scale.} In order to confirm the effect of the number of parameters of the foundation model constituting the backbone in the pretraining and fine-tuning strategy, we use the pretraining dataset and pretraining method from previous research that proposed successful foundation model construction\cite{wang2022advancing}. Specifically, the dataset, pretraining method, and foundation model structure are the same as in previous research, namely MillionAID\cite{long2021creating}, MAE\cite{he2022masked}, and vision transformer\cite{dosovitskiy2020image}. The only altered variable is the number of parameters constituting the model. By transforming the model to 86M, 605.26M, 1.3B, and 2.4B, we confirm that the performance of the foundation model improves in proportion to the number of parameters. {In addition, as a way to increase the number of parameters of the vision transformer for retaining the ability of object localization, it is suggested that not only the layer, hidden size, and dimension but also parallelism can be controlled.} {At last}, the performance of the foundation model is confirmed by fine-tuning on DOTA v2.0, DIOR-R, Potsdam\footnote{\url{https://www.isprs.org/education/benchmarks/UrbanSemLab/2d-sem-label-potsdam.aspx}} and LoveDA datasets, representing rotated object detection and semantic segmentation\cite{ding2021object, cheng2022anchor, wang2021loveda} {which require the ability of object localization.} To sum up, our work highlights the following contribution points:

\begin{itemize}
    \item We introduce the first foundation model tailored for remote sensing, featuring a billion-scale parameters. This significant advancement shows that larger models can greatly improve tasks like object detection and semantic segmentation in remote sensing, setting new standards for model complexity.
    \item Our work also unveils a effective method to enhance vision transformers through parallelism, making them suitable for the intricate needs of remote sensing. This scaling approach boosts the model's parameters and fine-tunes its performance for tasks demanding high-resolution insights and computational efficiency.
    \item Finally, our extensive experiments highlight the importance of pretraining on large datasets and how it influences task success. We find that larger models, especially those pretrained on comprehensive remote sensing datasets, exhibit better performance and sample efficiency in various downstream tasks, affirming the value of extensive pretraining and the positive impact of model size on performance.
\end{itemize}

\section{Related Works}\label{sect:MaterialData}

\subsection{Self-supervised Learning in Computer Vision}

\begin{figure}[t]
    \centering
    \includegraphics[width=\linewidth]{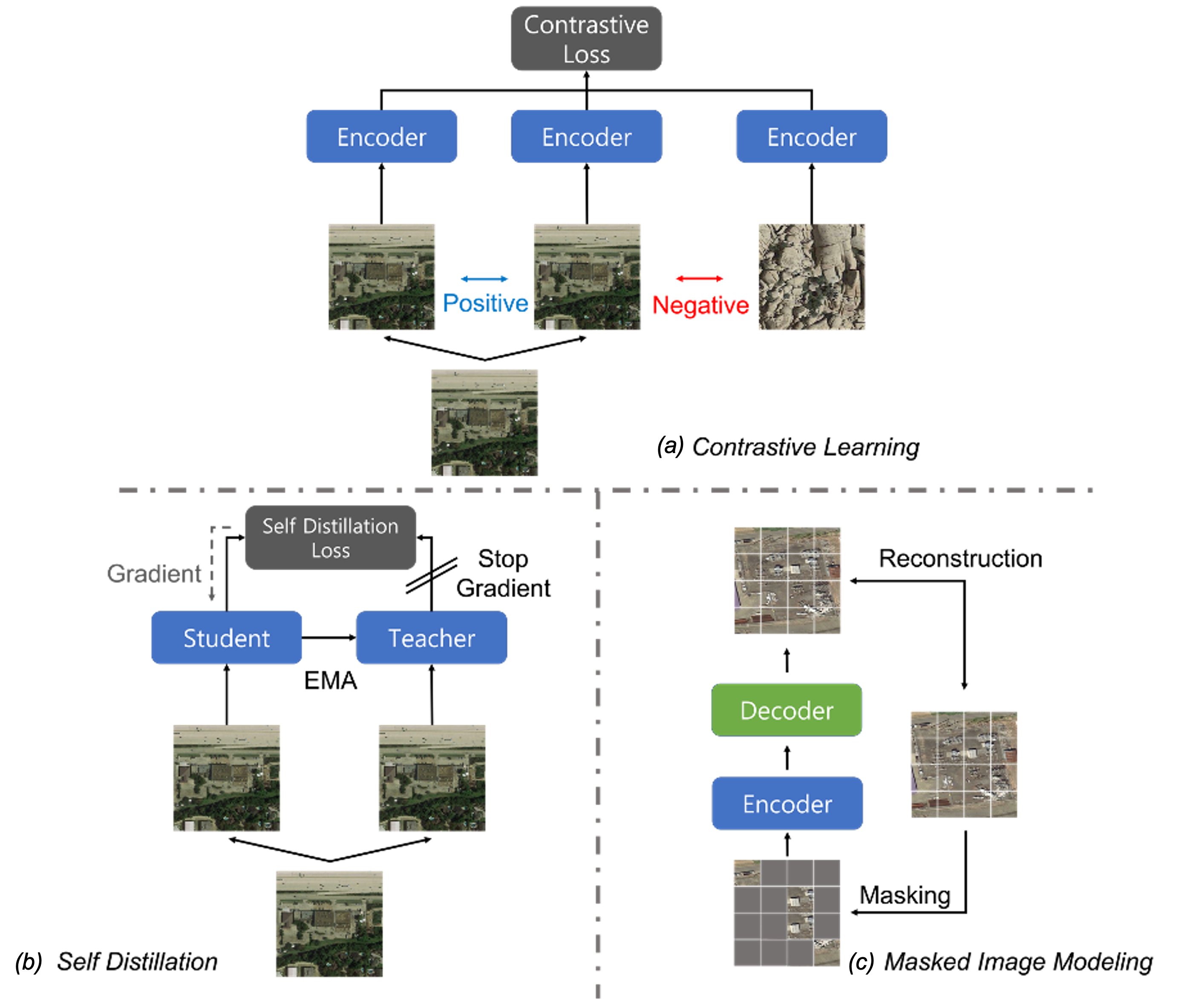}
    \caption{A brief introduction of self supervised learning, such as contrastive learning, self-distillation and masked image modeling in computer vision. (a) In contrastive learning, the positive pairs of data are brought closer together while the negative pairs are pushed further apart. (b) Self-distillation is a process of training a model to predict the relationships between multiple views of an unlabeled image. (c) Masked image modeling involves masking a portion of an image and then using a process to reconstruct the masked section.}
    \label{fig:selfsupcv}
\end{figure}

In recent years, the landscape of deep learning research in computer vision has experienced a significant shift towards the development and utilization of foundation models. Traditionally, the focus was predominantly on creating task-specific models, which demanded extensive resources for data collection, labeling, and training. This approach, while effective, was both time-consuming and costly. The advent of foundation models marks a paradigm shift, offering a more efficient and cost-effective solution to these challenges. Foundation models are designed to be versatile, pre-trained on large datasets of unlabeled data, and can be fine-tuned to a variety of tasks \cite{brown2020language, chowdhery2022palm, yuan2021florence}. This adaptability stems from their ability to extract meaningful representations from data, significantly reducing the resources required to develop task-specific models.

The creation of a foundation model hinges on three pivotal factors: a large-scale dataset, a robust model architecture, and an pretraining methodology. The domain of computer vision has witnessed the construction of extensive datasets, both single-modal and multi-modal, such as ImageNet \cite{deng2009imagenet}, JFT-300M \cite{sun2017revisiting}, LVIS \cite{gupta2019lvis}, and LAION-400M \cite{schuhmann2021laion}. Parallel to dataset development, there has been a concerted effort to expand the scale of model architectures. Inspired by the advancements in natural language processing models, which have reached up to 175B parameters, computer vision models have seen a significant increase in complexity, with models boasting 1.3B \cite{goyal2021self}, 2B \cite{zhai2022scaling}, and even 22B parameters \cite{dehghani2023scaling}. These large-scale models have demonstrated superior performance compared to their predecessors, underscoring the importance of scale in model architecture.

Pretraining methodologies are the linchpin in the development of foundation models. In the realm of computer vision, this often involves self-supervised learning techniques, which have shown great promise in leveraging unlabeled images to train models. As shown in \autoref{fig:selfsupcv}, this approach includes methods such as contrastive learning, self-distillation, and masked image modeling. Contrastive learning, for instance, utilizes unlabeled images to learn representations by discerning similarities and differences between image pairs \cite{chen2020simple, he2020momentum, oord2018representation}. Self-distillation further refines this concept by predicting relationships between multiple views of an image, optimizing the model through losses like MSE, InfoNCE, and knowledge distillation \cite{grill2020bootstrap, caron2021emerging, zhou2021ibot, chen2020improved}. Masked image modeling introduces a different angle by encouraging models to reconstruct parts of an image that have been intentionally obscured \cite{he2022masked, xie2022simmim, bao2021beit, peng2022beit}.
The methodologies for pretraining foundation models are diverse and largely dependent on the dataset modality in use. For single-modal learning, techniques such as SimCLR~\cite{chen2020simple}, BYOL~\cite{grill2020bootstrap}, DINO~\cite{caron2021emerging}, and iBOT~\cite{zhou2021ibot} have been employed, while multi-modal learning has benefited from approaches like CLIP~\cite{radford2021learning}, BEiT-3~\cite{wang2022image}, and Uni-Perceiver~\cite{li2022uniperceiver}.

\subsection{Foundation Models in Remote Sensing}
The exploration of foundation models in remote sensing is an emerging and vibrant field, drawing inspiration from significant advancements in self-supervised learning and the application of vision foundation model principles to remote sensing data. Remote sensing data, characterized by its integration of space-time coordinates and diverse spatial scales, provides a unique platform for applying and extending self-supervised learning techniques.
Foundation models in remote sensing leverage large-scale datasets to enhance model robustness and performance across a variety of tasks including object classification, detection, and semantic segmentation~\cite{cheng2017remote,bi2020multiple,bi2021local,wang2020looking,wang2024samrs}. These models benefit from pretraining on extensive datasets, allowing for quicker convergence and superior performance on downstream tasks compared to models trained from scratch. This approach mitigates overfitting on specific tasks by fostering rich feature learning during the pretraining phase \cite{tao2022self}.

A notable trend in this domain is the adaptation of single-modal and multi-modal contrastive learning methods. Single-modal learning, utilizing either instance-level or time-series-level signals, employs augmented perspectives of the same images as positive pairs and those from unrelated images as negative pairs. This technique is effectively applied using optical images in remote sensing, following methodologies such as SimCLR \cite{chen2020big, li2022global, bachman2019learning, tao2020remote, reed2022self}. Time-series-level contrastive learning, on the other hand, is designed for temporal robustness, utilizing datasets like the Functional Map of the World (fMoW)~\cite{christie2018functional} to handle spatial information with temporal variation \cite{ayush2021geography, manas2021seasonal}.
Multi-modal contrastive learning in remote sensing aims to harness rich feature representations through the use of diverse data modalities, including optical imagery, synthetic aperture radar, near-infrared, text, and audio. This approach includes SAR-optical contrastive learning, which benefits from the complementary nature of multi-modalities under various conditions \cite{jain2022multimodal, cha2021contrastive, yuan2023bridging, chen2021self, wang2022ssl4eo,9880533,9553741, huang2021qxs,guo2023skysense}. ~\cite{guo2023skysense} is a concurrent work described as a billion-scale multi-modal remote sensing foundation model, utilizing multi-modal contrastive learning to learn features across various modal and spatial granularities.
Furthermore, innovative methods like image-text and image-audio contrastive learning explore the relationship between remote sensing images and other data types, although these methods are less common due to the challenge of generating corresponding text or audio for remote sensing images \cite{mikriukov2022deep, heidler2023self, lu2017exploring, qu2016deep}.

Generative learning, including Masked Image Modeling (MIM)~\cite{he2022masked,xie2022simmim}, focuses on reconstructing an input image with parts of it obscured or removed, aiming to train a model on spatial or temporal reconstruction tasks. This method aligns with generative adversarial networks and auto-encoding techniques, facilitating the learning of features by reconstructing the corrupted areas of original images \cite{wang2022advancing, sun2022ringmo, cong2022satmae, reed2022scale, singh2018self, goodfellow2020generative, duan2018gan, yuan2022sits, yuan2020self,zhang2022consecutive,zhang2023object}.
Recent efforts in foundation models have also explored the amalgamation of contrastive learning and MIM strategies, alongside investigating multi-modal pretraining for both single- and multi-modal tasks using static imagery. For instance, these approaches utilize geo-location prediction and multiple views for self-supervised learning within established frameworks \cite{wang2022advancing, sun2022ringmo, ayush2021geography, manas2021seasonal, reed2022scale, wanyan2023dino, muhtar2023cmid, mendieta2023gfm, mall2023change, fuller2023croma, wang2023decur,han2024bridging}. These innovations demonstrate the potential of extending VFM techniques to space-time remote sensing data, contributing to the robust development of foundation models tailored for remote sensing applications.

In summary, the field of remote sensing is witnessing rapid advancements through the application of self-supervised learning methodologies, including single-modal and multi-modal contrastive learning and generative learning. These efforts are significantly enriching the feature learning capabilities of models in this domain, enhancing their applicability and performance across a wide range of remote sensing tasks.

\section{Methods}

In this section, we discuss the details of the model architecture (vision transformer) \cite{dosovitskiy2020image}, pretraining dataset (MillionAID) \cite{long2021creating}, and pretraining method (MAE)\cite{he2022masked}. Additionally, we explain how to properly scale up the vision transformer, which is one of the essential contributions of this work\cite{touvron2022three}. Furthermore, we address how to adapt the plain vision transformer for tasks like rotated object detection and semantic segmentation, which require high-resolution input imagery and object localization ability\cite{li2022exploring}.

\subsection{Self Supervised Learning by MAE}

To demonstrate that the performance of the foundation model improves with the increase in the number of model parameters when pretrained using the same number of datasets in the remote sensing field, we pretrain models with different numbers of parameters using MAE\cite{he2022masked} and the large-scale remote sensing imagery dataset, MillionAID\cite{long2021creating}.

\subsubsection{MillionAID}

The MillionAID dataset\cite{long2021creating} is easily accessible and contains the largest number of RGB images. This dataset is designed for scene classification tasks and includes approximately 1,000,848 non-overlapping remote sensing scenes. It features 51 detailed classes, organized in a three-level tree structure.
The 51 individual components (leaf nodes) are assembled into 28 broader categories (parent nodes) at the secondary tier, which are subsequently classified into eight principal clusters (nodes) at the primary level. These dominant clusters embody diverse land use classifications, including agricultural, industrial, public amenities, and residential zones.
To build a robust model at varying ground sample distances (GSD), the images are collected from Google Earth, with resolutions ranging from 0.5m to 153m per pixel. The images also vary in size, from 110 $\times$ 110 to 31,672 $\times$ 31,672 pixels. While each image is mapped to a class, these images are considered unlabeled data, so only the images are used for pretraining. Although it is evident that more pretraining images lead to better foundation model performance, this paper focuses on the effect of the number of model parameters during training, using only MillionAID for a fair comparison with other research\cite{wang2022advancing}.

\subsubsection{MAE}
The Masked Autoencoder (MAE)~\cite{he2022masked} for self-supervised learning utilizes a novel approach of reconstructing the masked regions of an input image. Let us denote an image \( I \) that is divided into \( N \) non-overlapping patches, \( \{p_1, p_2, ..., p_N\} \). Each patch \( p_i \) is embedded into a latent space using a linear projection to obtain a corresponding patch embedding \( x_i \).
The core of MAE lies in its masking strategy, where a random subset of these embeddings, say \( M \) out of \( N \) (typically \( M = 0.75N \)), are masked out, resulting in a reduced set of visible patches \( V \). The visible patches are represented as \( V = \{x_{i_1}, x_{i_2}, ..., x_{i_{N-M}}\} \), where \( i_k \) denotes the indices of the unmasked patches.

The encoder processes only the visible patches \( V \), generating a set of latent representations \( Z = \{z_{i_1}, z_{i_2}, ..., z_{i_{N-M}}\} \). These representations are then fed into a decoder, alongside positional embeddings of the masked patches. The decoder aims to reconstruct the original pixel values of the masked patches, \( \hat{p}_{j_1}, \hat{p}_{j_2}, ..., \hat{p}_{j_M} \), where \( j_k \) denotes the indices of the masked patches.
The training objective of MAE is to minimize the reconstruction error, which is quantified by the mean squared error (MSE) between the reconstructed patches and the original patches:
\[
\mathcal{L} = \frac{1}{M} \sum_{j=1}^{M} \| \hat{p}_{j} - p_{j} \|^2 .
\]
This loss function ensures that the model learns to accurately predict the pixel values of the masked regions, leveraging only the information available from the visible regions. The effectiveness of this pretraining strategy is demonstrated by its ability to enhance downstream task performance, suggesting that the encoder captures useful, high-level representations.

\subsubsection{Scaling Up Vision Transformer}

\begin{figure}[t]
    \centering
    \includegraphics[width=\linewidth]{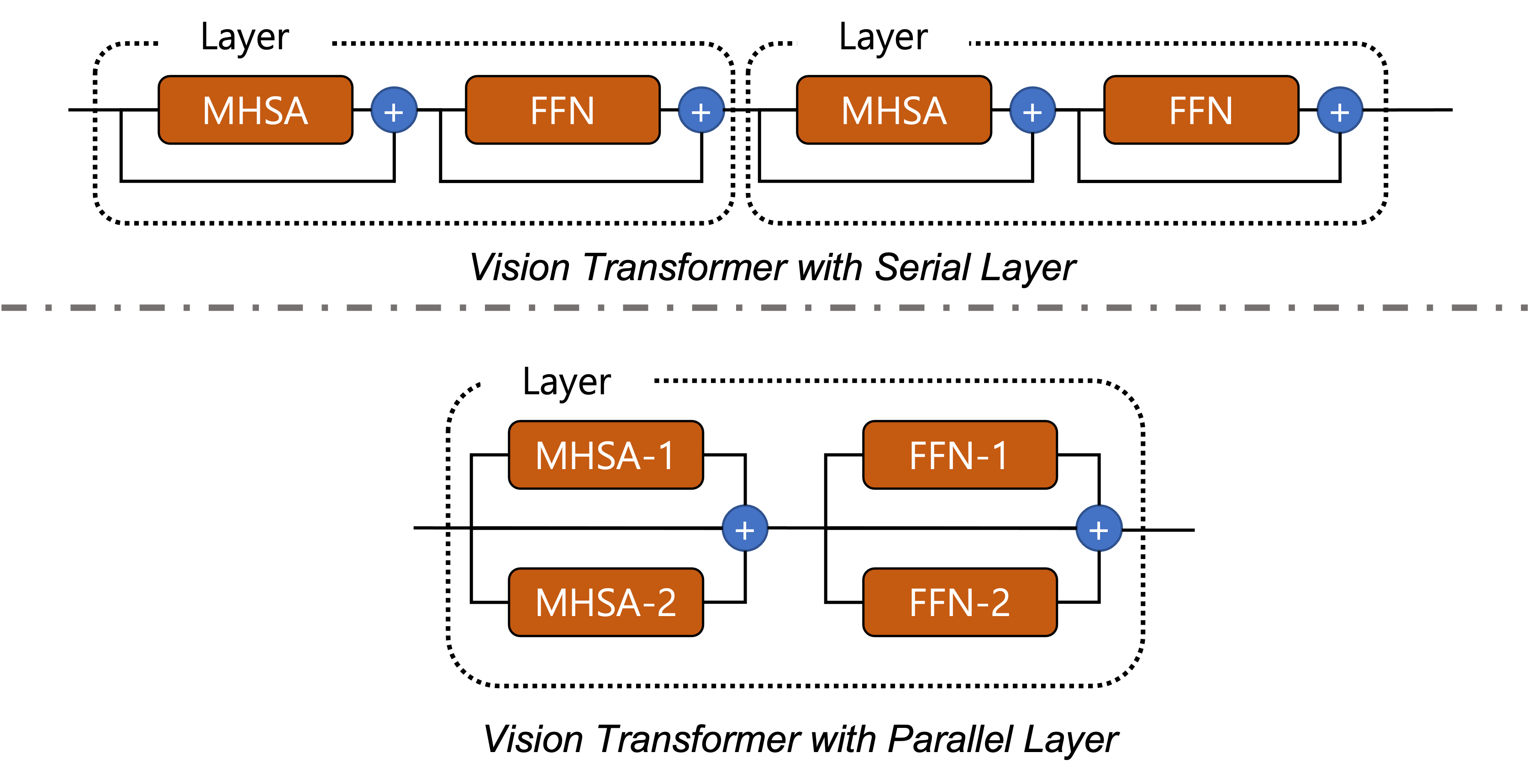}
    \caption{This figure explains how to effectively increase the number of parameters of the vision transformer, and the two models have substantially the same amount of computation and number of parameters. In the field of natural language processing, multi head self attention and feed forward blocks are configured only serially, but there is the difference of performance even if they are configured in parallel. Like 12 layers with 1 parallelism and 6 layers with 2 parallelism, if the same number is obtained when multiplying the layer and parallelism, the backbone has the same number of parameters and the same flops.}
    \label{fig:parallel vision transformer}
\end{figure}

While several rules exist for scaling up the transformer structure in natural language processing, which heavily relies on transformer structures\cite{devlin2018bert, radford2019language}, these rules may not be as effective for scaling up vision transformers. For instance, in most self-supervised learning papers using vision transformers in computer vision, only classification and semantic segmentation performance are introduced, and it is challenging to find object detection performance\cite{caron2021emerging,he2022masked,peng2022beit, chen2021empirical}. If object detection performance is presented, the research typically involves a model with a significantly smaller number of parameters\cite{zhou2021ibot}. As shown in \autoref{fig:parallel vision transformer}, a study connects the multi-head self-attention structure and feed-forward blocks in parallel, which has a similar effect as stacking them serially\cite{touvron2022three}. Based on this approach, this paper demonstrates that object detection and semantic segmentation, which are crucial downstream tasks in remote sensing, can be performed effectively by scaling up the vision transformer in parallel.

Traditional scaling methods in natural language processing primarily extend the depth of the transformer architecture by serially stacking multi-headed self-attention (MHSA) and feed-forward network (FFN) layers. However, for Vision Transformers (ViTs), we explore a parallel configuration to enhance performance for vision-related tasks, particularly those requiring fine-grained spatial understanding such as object detection and segmentation.
Given a ViT with sequential layers, the output of the $l^{th}$ layer, $x_{l+1}$, is typically computed as follows:
\begin{equation}
x_{l+1} = \text{FFN}_l(\text{MHSA}_l(x_l) + x_l) + \text{MHSA}_l(x_l) + x_l.
\end{equation}
In a parallel configuration, we propose the simultaneous processing of MHSA and FFN blocks. Thus, the outputs for two consecutive layers are reformulated as:
\begin{align}
x_{l+1} &= x_l + \text{MHSA}_l(x_l) + \text{MHSA}_{l+1}(x_l), \\
x_{l+2} &= x_{l+1} + \text{FFN}_l(x_{l+1}) + \text{FFN}_{l+1}(x_{l+1}).
\end{align}
This parallel structure hypothesizes that as the network depth increases, the incremental effect of each layer diminishes. This implies that the approximation $r(x + r'(x)) \approx r(x)$, where $r$ and $r'$ represent different residual blocks, becomes increasingly valid. By leveraging this principle, the parallel configuration aims to maintain the computational efficiency while potentially simplifying optimization and improving the model's performance on complex visual tasks.

\subsubsection{Implementation Detail for Pretraining}

\begin{figure*}[t]
    \centering
    \includegraphics[width=\linewidth]{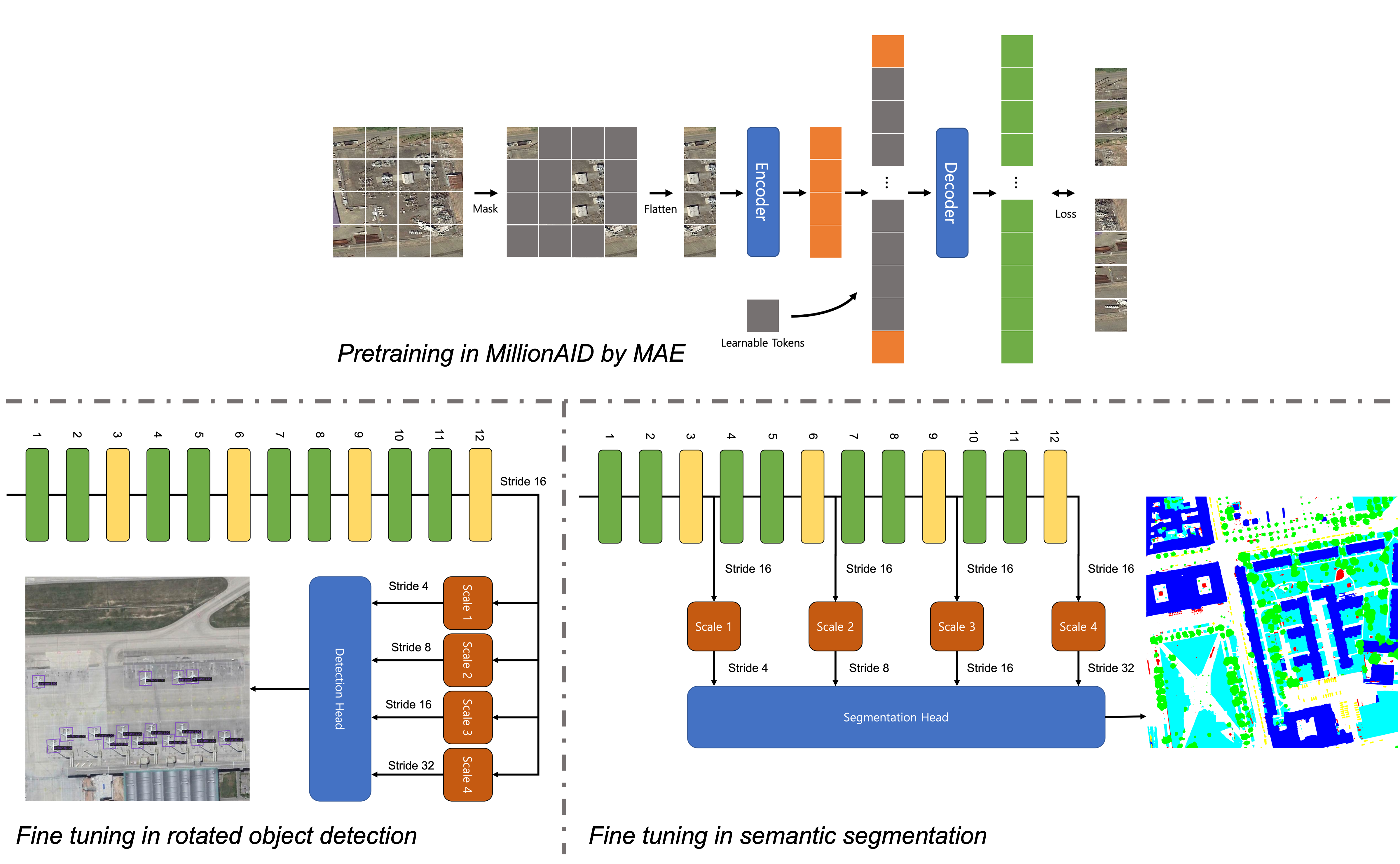}
    \caption{This figure shows overall flows for pretraining and downstream tasks. The plain vision transformer is pretrained by MAE with remote sensing imagery dataset, MillionAID. Then, plain vision transformer is converted to ViTDET structure with local and global attention for downstream tasks which is rotated object detection and semantic segmentation. In order to upsample and downsample features, scale blocks are adopted after ViTDET backbone. The scale block 1 consists of serially connected transposed convolution, normalization, GELU, and transposed convolution\cite{hendrycks2016gaussian, long2015fully}. The scale block 2 is only transposed convolution. The scale block 3 is identity block. The scale block 4 is max pooling with kernel size 2. All transposed convolutions used in scale block is with kernel size 2 and stride size 2.}
    \label{fig:ViTDET}
\end{figure*}

This paper primarily focuses on analyzing the scaling up of the model. Therefore, the configuration of the models used in this study is introduced first. The detailed information of the vision transformer for the experiment is shown in \autoref{tab:model configuration}. In order to scale up the vision transformer effectively, the number of parallelism, hidden size, MLP size, and heads are changed, while the number of layers is fixed at 12. {The method of naming the model is as follows. The model name is follows ViT-(A)(B)$\times$(C). (A), (B) and (C) stand for hidden size, layers and parallelism, respectively. For example, ViT-G12$\times$4 has the 2,048 hidden size, 12 layers, and 4 parallelism. In case of ViT-B12$\times$1, this model has same structure of standard vision transformer base.}
During pretraining, input images are resized and cropped to 224 x 224 using random resized crop transformation\footnote{\url{https://pytorch.org/vision/main/generated/torchvision.transforms.RandomResizedCrop.html}}. After resizing and random cropping, horizontal flip and vertical flip are sequentially applied with a 50\% probability. The input image is then divided into $16 \times 16$ patches before being fed into the vision transformer encoder, resulting in $(224 / 16) \times (224 / 16)$ sampled patches. In the original MAE, pretraining is applied with 1,600 epochs \cite{he2022masked}. However, since models with a large number of parameters can experience overfitting to the pretext task~\cite{wang2022rethinking,wang2024memorization,meehan2023ssl} (MAE in this paper), the models are pretrained with 400 epochs using the AdamW optimizer\cite{loshchilov2017decoupled} and a batch size of 2,048. {In general, the learning rate should be different according to the batch size even in the same experiment. Since it is not appropriate for expressing the hyper parameters, the learning rate is expressed with the value of base learning rate. The effective learning rate is as follows, $\text{batch size} \times \text{BaseLR} / 256$.} It is worth noting that only the ViT-G12x4 model has a different base learning rate than the other models. This is because the loss did not converge when a learning rate of $1.5e^{-4}$ was applied, but it did converge when 1e-4 was used. For the mask portion, the same 75\% value is used as in previous research\cite{wang2022advancing}. Additionally, due to the large size of the model presented in this paper, activation checkpointing and fp16 were applied for pretraining to address the issue of insufficient GPU memory~\cite{rajbhandari2020zero}.

\begin{table}[t]
     \centering
     \caption{the configuration of models handeled in this research.} %
     \begin{tabular}{l|c|c|c|c}
        \hline
         Name & Hidden size & MLP size  & Heads & Params \\ \hline
         ViT-B12$\times$1\cite{wang2022advancing} & 768 & 3072 & 12 & 86M \\ \hline
         ViT-L12$\times$4 & 1024 & 4096 & 16 & 605.26M \\ \hline
         ViT-H12$\times$4 & 1536 & 6144 & 16 & 1.36B \\ \hline
         ViT-G12$\times$4 & 2048 & 8192 & 32 & 2.42B \\ \hline
     \end{tabular}
     \label{tab:model configuration}
 \end{table}

\subsection{Fine Tuning Vision Transformer for Object Localization} \label{subsec:vitdet}

 \begin{table}[t]
     \centering
     \caption{the hyperparameters of pretraining.} %
     \begin{tabular}{l|c|l|c}
        \hline
         Name & Epochs & Base LR  & Weight Decay \\ \hline
         ViT-B12$\times$1\cite{wang2022advancing} & 1600 & 0.00015 & 0.05 \\ \hline
         ViT-L12$\times$4 & 400 & 0.00015 & 0.05 \\ \hline
         ViT-H12$\times$4 & 400 & 0.00015 & 0.05 \\ \hline
         ViT-G12$\times$4 & 400 & 0.0001 & 0.05 \\ \hline
     \end{tabular}
     \label{tab:pretraining detail}
 \end{table}
 
Downstream tasks such as rotated object detection and semantic segmentation require the ability to localize objects and high-resolution input images, e.g., $1024 \times 1024$. However, the pretrained backbone has a plain and non-hierarchical structure, which calculates the relationship between all patches. Consequently, it is not suitable to use without transformation in terms of operation speed and GPU memory. To address this problem, local and global attention introduced in ViTDET\cite{li2022exploring} is applied, which can reduce computation and GPU memory usage without degrading performance. As shown in \autoref{fig:ViTDET}, local attention with a window size of 14 is applied to blocks 1, 2, 4, 5, 7, 8, 10, and 11, while global attention is applied to blocks 3, 6, 9, and 12. In rotated object detection tasks, the feature of the last layer is resampled to have a ratio of 4, 2, 1, 0.5 through each scale block, known as a simple pyramid network\cite{li2022exploring}. In semantic segmentation, the features of layers 3, 6, 9, and 12 are resampled to have ratios of 4, 2, 1, and 0.5 through each scale block, respectively. These features are then fed into the detection head and segmentation head.

\section{Experimental Results}

In this section, we evaluate the performance of the proposed models on remote sensing downstream tasks, including rotated object detection and semantic segmentation, after pretraining. We then examine whether the performance increases in proportion to the number of parameters of the pretrained model across all datasets and benchmarks. We compare the performance with other studies, but the comparison group pretrained in remote sensing mainly includes the previous studies \cite{wang2022advancing}. This is because the amount of pretraining data (MillionAID) and the methodology used in pretraining (MAE) should be the same for fair comparison.

To demonstrate whether the ability to extract representations increases as the number of parameters included in the model increases, we conduct sample efficiency experiments with DIOR-R for rotated object detection and Potsdam for semantic segmentation. To obtain sample efficiency results, we use 1\%, 5\%, 10\%, 50\%, and 100\% of the training dataset and 100\% of the test dataset for evaluation.

\subsection{Rotated Object Detection}

\subsubsection{Dataset}

In this paper, two benchmark dataset is adopted to evaluate the performance of pretrained models: DOTA v2.0\cite{ding2021object} and DIOR-R\cite{cheng2022anchor}.

\textbf{DOTA v2.0}. The DOTA dataset is the most well-known dataset as a benchmark for rotated object detection in the remote sensing field. We use DOTA v2.0 for evaluation because of its better integrity than v1.0. To make the model robust at various resolutions, images are collected from multiple sources and consist of images with pixel ranges from 800 $\times$ 800 to 20,000 $\times$ 20,000. DOTA v2.0 has 18 categories including harbor, swimming-pool, roundabout, bridge, baseball-diamond, ground-track-field, soccer-ball-field, storage-tank, basketball-court, large-vehicle, plane, tennis-court, ship, helicopter, helipad, container-crane, airport, and small-vehicle. Also, it has 11,268 images and 1,793,658 instances. The training dataset contains 1,830 images and 268,627 instances. The validation dataset contains 593 images and 81,048 instances. And, the test-dev contains 2,792 images and 353,346 instances. However, the test-dev is published with only images. The test-challenge dataset contains 6,053 images and 1,090,637 instances. However, because the test-challenge dataset is only published during challenge, it can not be accessible now. We evaluate our model by submitting the inference result to the server\footnote{\url{https://captain-whu.github.io/DOTA/evaluation.html}}.

\textbf{DIOR-R}.  The DIOR-R dataset is also widely used as a benchmark for evaluating rotated object detection performance in remote sensing. The DIOR-R dataset is divided into training, validation and test. It has 20 categories including expressway-toll-station, chimney, baseball-field, vehicle, harbor, basketball-court, golf-field, tennis-court, storage-tank, windmill, train-station, bridge, ground-track-field, ship, airport, airplane, Expressway-Service-area, dam, stadium, and overpass. It includes 11,738 images and 68,073 instances by bundling training with validation. The test contains 11,738 images and consists of 124,445 instances. All images are 800x800 in size, pixel resolution is from 0.5m to 30m. Unlike DOTA v2.0, the training, validation and test datasets are totally published with both images and instances. We evaluate our model locally using the published dataset.

\subsubsection{Implementation Details and Experiment Settings}

As in other research, we train the rotated object detection models with the same hyperparameters, regardless of the dataset, and only change the pretrained backbone. The detection models are trained based on the mmrotated framework\footnote{\url{https://github.com/open-mmlab/mmrotate}} with added dataset functionality \cite{zhou2022mmrotate}. We use the AdamW optimizer \cite{loshchilov2017decoupled}, with a learning rate of 0.0001 and weight decay of 0.05. The training schedule is designed for a batch size of 2 and 12 epochs, with the learning rate being reduced by 10$\times$ at the 8th and 11th epochs. We also apply a 0.8 layer-wise learning rate decay strategy and a drop path rate of 0.1 to ViTDET.

We select and fix the roi transformer \cite{ding2019learning} as the detection head, which has the best performance in the mmrotate framework. Since \cite{wang2022advancing} has evaluated the performance with Oriented R-CNN \cite{xie2021oriented} but not on the DOTA v2.0 dataset, we re-implement and evaluate the detection model composed of the backbone weight published by \cite{wang2022advancing} and the roi transformer. The backbone is converted into the ViTDET architecture, as shown in \autoref{subsec:vitdet}.

Due to the wide range of pixel sizes in the DOTA v2.0 dataset, we crop each image to a size of 1024 $\times$ 1024 with a stride of 824. For the DIOR-R dataset, images are already set to a constant size of 800$\times$800, so cropping is not performed. During training, we use only random horizontal and vertical flips as data augmentation for both datasets. Test-time augmentation is not applied during inference on test images, and models are evaluated based on average precision (AP) for each category and mean average precision (mAP), which is the average of APs across all categories.

\begin{figure*}[p!bt]{}
    \centering
    \includegraphics[width=0.9\textwidth]{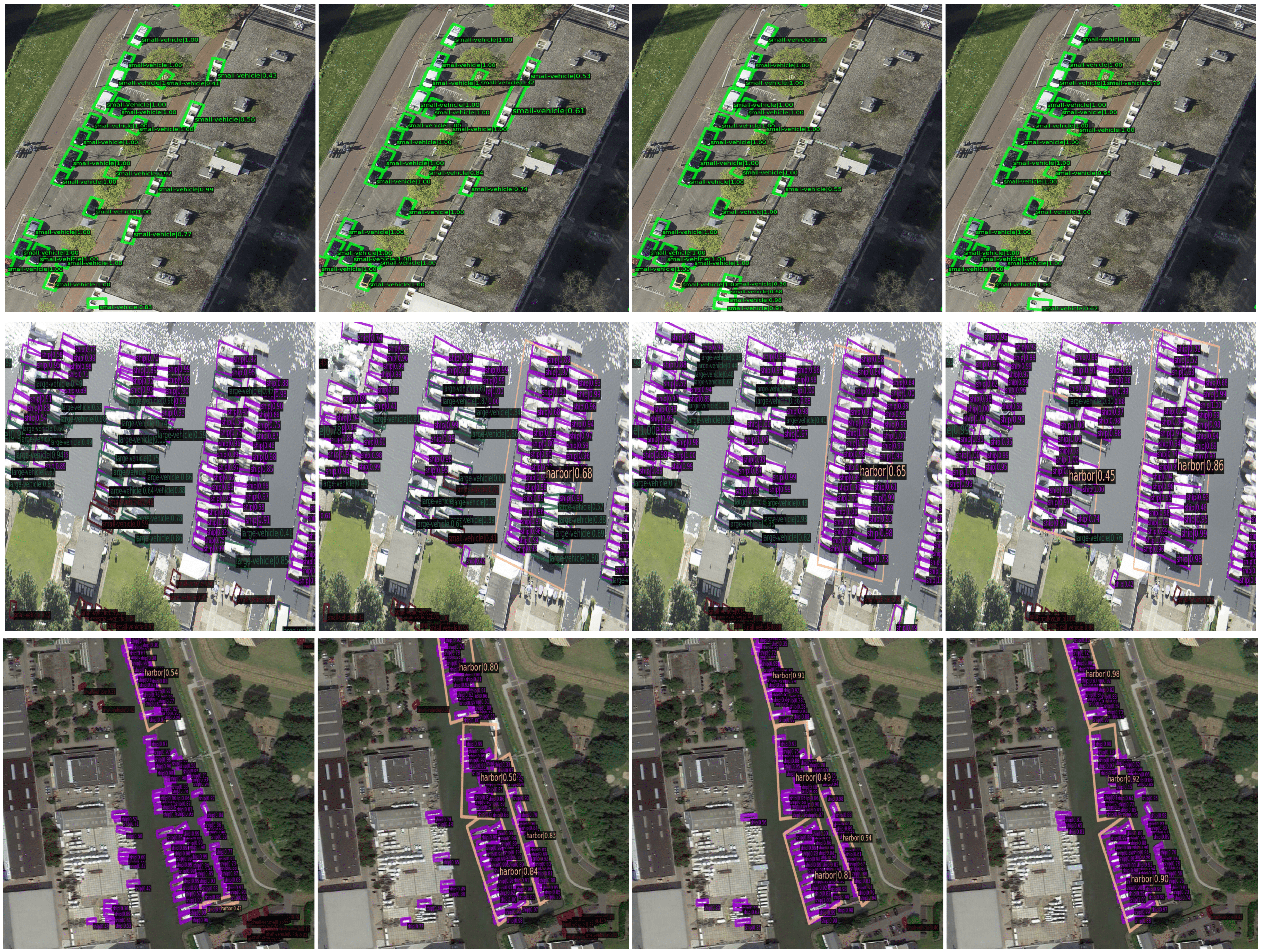}
    \includegraphics[width=0.9\textwidth]{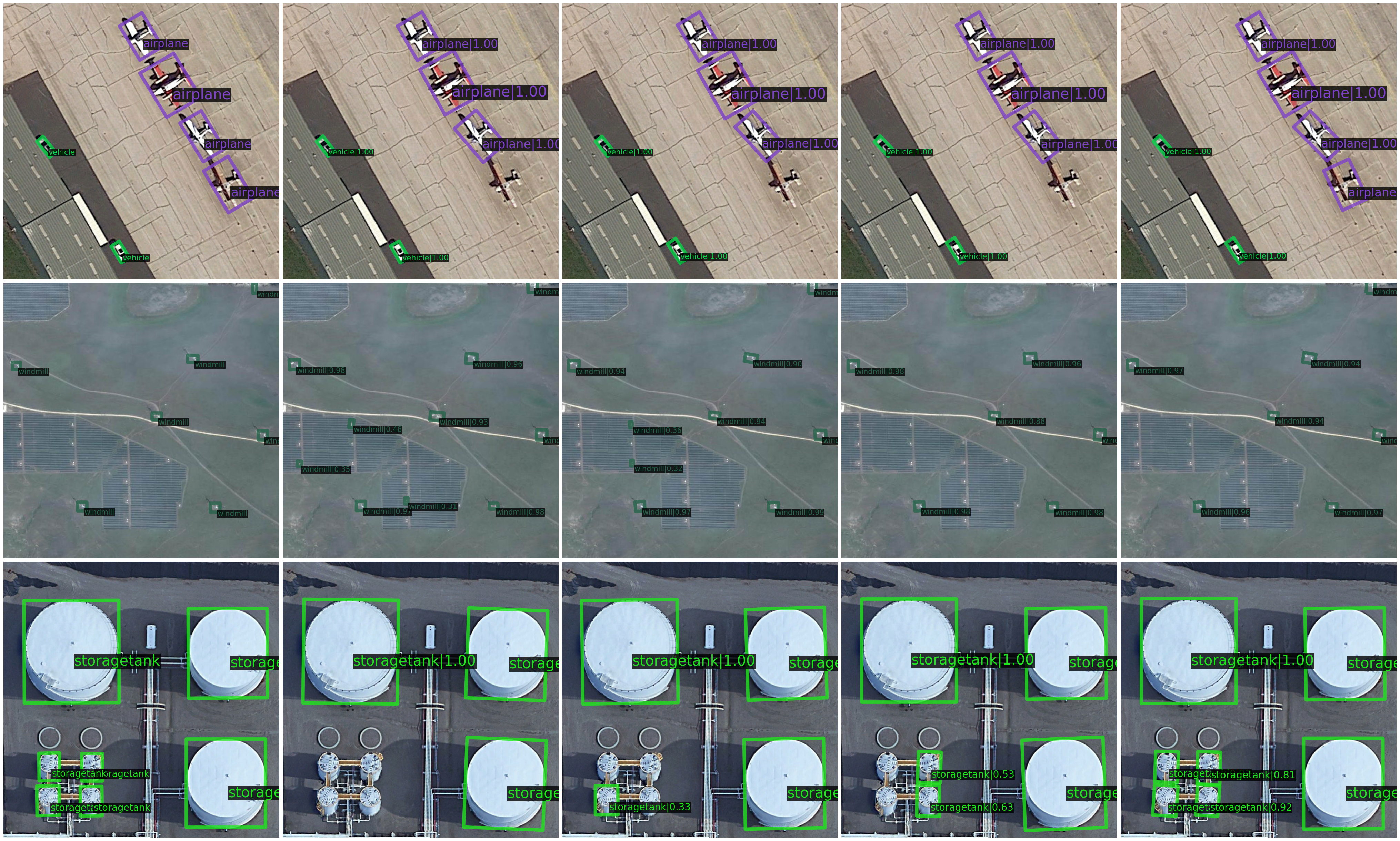}
    \caption{Visualization results of the proposed model. The first through third rows are the results of the DOTA v2.0 dataset. Since the label of test dataset in DOTA v2.0 is unavailable, the images from left to right are ViT-B12$\times$1, ViT-L12$\times$4, ViT-H12$\times$4, and ViT-G12$\times$4. The fourth to sixth rows are the results of the DIOR-R dataset. The images from left to right are label, ViT-B12$\times$1, ViT-L12$\times$4, ViT-H12$\times$4, and ViT-G12$\times$4. }
    \label{fig:vis_rotated_od}
\end{figure*}

\subsubsection{Experiment Results}

\begin{table*}[ht]{\textwidth=0mm}
    \centering
    \caption{the results of class-wise AP and mAP on DOTA v2.0. Because existing benchmarks have not been performed on mmrotate framework, the $\dagger$ means the result of evaluation using RoI Transformer in Res50 with mmrotate framework. $\diamondsuit$ is the result re-implemented in mmrotate framework using the vision transformer weight published by \cite{wang2022advancing} with RoI Transformer.} %
    \setlength{\tabcolsep}{5pt}
    {\tiny
    \begin{tabular}{l|l|c c c c c c c c c c c c c c c c c c | c}
    \hline
       Backbone & Method & PL & BD & BR & GTF & SV & LV & SH & TC & BC & ST & SBF & RA & HA & SP & HC & CC & AP & HL & mAP  \\ \hline
       
       Res50(IMP)\cite{he2016deep} & RetinaNet\cite{lin2017focal} & 70.63 & 47.26 & 39.12 & 55.02 & 38.1 & 40.52 & 47.16 & 77.74 & 56.86 & 52.12 & 37.22 & 51.75 & 44.15 & 53.19 & 51.06 & 6.58 & 64.28 & 7.45 & 46.68 \\
       Res50(IMP)\cite{he2016deep} & MR\cite{he2017mask}& 76.2 & 49.91 & 41.61 & 60 & 41.08 & 50.77 & 56.24 & 78.01 & 55.85 & 57.48 & 36.62 & 51.67 & 47.39 & 55.79 & 59.06 & 3.64 & 60.26 & 8.95 & 49.47 \\
       Res50(IMP)\cite{he2016deep} & CMR\cite{chen2019hybrid} & 77.01 & 47.54 & 41.79 & 58.2 & 41.58 & 51.74 & 57.86 & 78.2 & 56.75 & 58.5 & 37.89 & 51.23 & 49.38 & 55.98 & 54.59 & 12.31 & 67.33 & 3.01 & 50.04 \\
       Res50(IMP)\cite{he2016deep} & HTC\cite{chen2019hybrid} & 77.69 & 47.25 & 41.15 & 60.71 & 41.77 & 52.79 & 58.87 & 78.74 & 55.22 & 58.49 & 38.57 & 52.48 & 49.58 & 56.18 & 54.09 & 4.2 & 66.38 & 11.92 & 50.34 \\
       Res50(IMP)\cite{he2016deep} & FR OBB\cite{xia2018dota} & 71.61 & 47.2 & 39.28 & 58.7 & 35.55 & 48.88 & 51.51 & 78.97 & 58.36 & 58.55 & 36.11 & 51.73 & 43.57 & 55.33 & 57.07 & 3.51 & 52.94 & 2.79 & 47.31
       \\
       Res50(IMP)\cite{he2016deep} & FR OBB + Dp\cite{dai2017deformable} & 71.55 & 49.74 & 40.34 & 60.4 & 40.74 & 50.67 & 56.58 & 79.03 & 58.22 & 58.24 & 34.73 & 51.95 & 44.33 & 55.1 & 53.14 & 7.21 & 59.53 & 6.38 & 48.77 \\
       Res50(IMP)\cite{he2016deep} & FR H-OBB\cite{xia2018dota} & 71.39 & 47.59 & 39.82 & 59.01 & 41.51 & 49.88 & 57.17 & 78.36 & 56.87 & 58.24 & 37.66 & 51.86 & 44.61 & 55.49 & 54.74 & 7.56 & 61.88 & 6.6 & 48.9 \\
       Res50(IMP)\cite{he2016deep} & FROBB + RT\cite{ding2019learning} & 71.81 & 48.39 & 45.88 & 64.02 & 42.09 & 54.39 & 59.92 & \textbf{\textcolor{blue}{82.7}} & 63.29 & 58.71 & 41.04 & 52.82 & 53.32 & 56.18 & 57.94 & 25.71 & 63.72 & 8.7 & 52.81 \\
       Res50(IMP)$\dagger$ \cite{he2016deep} & FROBB + RT\cite{ding2019learning} & 77.84 & 51.54 & 45.97 & \textbf{\textcolor{red}{65.78}} & 43.25 & 55.03 & 60.38 & 79.45 & 62.98 & 59.85 & \textbf{\textcolor{red}{46.89}} & 56.53 & 54.06 & 56.71 & 51.65 & 21.31 & 68.05 & 8.32 & 53.64 \\ \hline
       ViT-B12$\times$1(MAE)$\diamondsuit$ \cite{wang2022advancing} & FROBB + RT\cite{ding2019learning} & 79.49 & \textbf{\textcolor{red}{55.86}} & \textbf{\textcolor{red}{50.12}} & 65.36 & 43.82 & 56.63 & 61.18 & 79.07 & 62.24 & 60.62 & 41.71 & 57.88 & \textbf{\textcolor{red}{58.48}} & \textbf{\textcolor{red}{64.84}} & 58.83 & \textbf{\textcolor{red}{35.13}} & \textbf{\textcolor{red}{89.41}} & \textbf{\textcolor{blue}{14.14}} & 57.49 \\
       ViT-L12$\times$4(MAE)(Ours) & FROBB + RT\cite{ding2019learning} & 79.75 & 53.34 & 49.02 & 65.18 & 43.8 & \textbf{\textcolor{blue}{56.83}} & \textbf{\textcolor{blue}{61.28}} & \textbf{\textcolor{red}{83.08}} & 60.77 & \textbf{\textcolor{red}{66.83}} & 42.33 & \textbf{\textcolor{red}{58.33}} & 57.39 & \textbf{\textcolor{red}{64.84}} & 67.31 & 30.81 & 80.69 & 13.8 & 57.52 \\
       ViT-H12$\times$4(MAE)(Ours) & FROBB + RT\cite{ding2019learning} & \textbf{\textcolor{blue}{79.8}} & 53.26 & 49.32 & 64.28 & \textbf{\textcolor{blue}{43.9}} & 56.46 & 61.18 & 78.98 & \textbf{\textcolor{blue}{63.53}} & \textbf{\textcolor{blue}{60.71}} & 42.76 & 57.74 & 57.84 & 64.43 & \textbf{\textcolor{blue}{67.34}} & \textbf{\textcolor{blue}{34.69}} & 89.35 & \textbf{\textcolor{red}{14.3}} & \textbf{\textcolor{blue}{57.77}} \\
       ViT-G12$\times$4(MAE)(Ours) & FROBB + RT\cite{ding2019learning} & \textbf{\textcolor{red}{80.12}} & \textbf{\textcolor{blue}{54.12}} & \textbf{\textcolor{blue}{50.07}} & \textbf{\textcolor{blue}{65.68}} & \textbf{\textcolor{red}{43.98}} & \textbf{\textcolor{red}{60.07}} & \textbf{\textcolor{red}{67.85}} & 79.11 & \textbf{\textcolor{red}{64.38}} & 60.56 & \textbf{\textcolor{blue}{45.98}} & \textbf{\textcolor{blue}{58.26}} & \textbf{\textcolor{blue}{58.31}} & \textbf{\textcolor{blue}{64.82}} & \textbf{\textcolor{red}{69.84}} & 32.78 & \textbf{\textcolor{blue}{89.37}} & 11.07 & \textbf{\textcolor{red}{58.69}} \\ \hline
    \end{tabular}
    }
    \label{tab:dota2 table}
\end{table*}

\begin{table*}[ht]{\textwidth=0mm}
    \centering
    \caption{the results of class-wise AP and mAP on DIOR-R. As same with \autoref{tab:dota2 table}, $\diamondsuit$ is the result re-implemented in mmrotate framework using the vision transformer weight published by \cite{wang2022advancing} with RoI Transformer.} %
    \setlength{\tabcolsep}{3.75pt}
    {\tiny
    \begin{tabular}{l|l|c c c c c c c c c c c c c c c c c c c c | c }
        \hline
       Backbone & Method & APL & APO & BF & BC & BR & CH & DAM & ETS & ESA & GF & GTF & HA & OP & SH & STA & STO & TC & TS & VE & WM & mAP \\ \hline

       Res50(IMP)\cite{he2016deep} & RetinaNet\cite{lin2017focal} & 61.49 & 28.52 & 73.57 & 81.17 & 23.98 & 72.54 & 19.94 & 72.39 & 58.2 & 69.25 & 79.54 & 32.14 & 44.87 & 77.71 & 67.57 & 61.09 & 81.46 & 47.33 & 38.01 & 60.24 & 57.55 \\

       Res50(IMP)\cite{he2016deep} & FR OBB\cite{xia2018dota} & 62.79 & 26.8 & 71.72 & 80.91 & 34.2 & 72.57 & 18.95 & 66.45 & 65.75 & 66.63 & 79.24 & 34.95 & 48.79 & 81.14 & 64.34 & 71.21 & 81.44 & 47.31 & 50.46 & 65.21 & 59.54 \\

       Res50(IMP)\cite{he2016deep} & FROBB + RT\cite{ding2019learning} & 63.34 & 37.88 & 71.78 & 87.53 & 40.68 & 72.6 & 26.86 & 78.71 & 68.09 & 68.96 & 82.74 & 47.71 & 55.61 & 81.21 & 78.23 & 70.26 & 81.61 & 54.86 & 43.27 & 65.52 & 63.87 \\

       Res50(IMP)\cite{he2016deep} & Gliding Vertex\cite{xu2020gliding} & 65.35 & 28.87 & 74.96 & 81.33 & 33.88 & 74.31 & 19.58 & 70.72 & 64.7 & 72.3 & 78.68 & 37.22 & 49.64 & 80.22 & 69.26 & 61.13 & 81.49 & 44.76 & 47.71 & 65.04 & 60.06 \\

       Res50(IMP)\cite{he2016deep} & AOPG\cite{cheng2022anchor} & 62.39 & 37.79 & 71.62 & 87.63 & 40.9 & 72.47 & 31.08 & 65.42 & 77.99 & 73.2 & 81.94 & 42.32 & 54.45 & 81.17 & 72.69 & 71.31 & 81.49 & 60.04 & \textbf{\textcolor{red}{52.38}} & 69.99 & 64.41 \\
       
       Res50(IMP)\cite{he2016deep} & DODet\cite{cheng2022dual} & 63.4 & 43.35 & 72.11 & 81.32 & 43.12 & 72.59 & 33.32 & 78.77 & 70.84 & 74.15 & 75.47 & 48 & 59.31 & 85.41 & 74.04 & 71.56 & 81.52 & 55.47 & 51.86 & 66.4 & 65.1 \\ \hline

       ViT-B(MAE)\cite{wang2022advancing} & Oriented R-CNN\cite{xie2021oriented} & 81.04 & 41.86 & 80.79 & 81.39 & 44.83 & 78.35 & 35.12 & 67.67 & 84.85 & 75.44 & 80.8 & 37.65 & 59.33 & 81.15 & 78.7 & 62.87 & 89.83 & 56.17 & 49.87 & 65.36 & 66.65 \\

       ViTAE-B(MAE)\cite{wang2022advancing} & Oriented R-CNN\cite{xie2021oriented} & 81.3 & 46.73 & 81.09 & 87.62 & 47.38 & 79.79 & 31.99 & 69.72 & 86.71 & 76.23 & 82.13 & 42.47 & 60.45 & 81.2 & 80.11 & 62.75 & 89.75 & 64.56 & 50.77 & 65.33 & 68.4 \\

       ViT-B + VSA(MAE)\cite{wang2022advancing} & Oriented R-CNN\cite{xie2021oriented} & 81.2 & 50.68 & 80.95 & 87.41 & 51.27 & \textbf{\textcolor{red}{80.87}} & 34.61 & 76.4 & 88.32 & 78.21 & 83.31 & 45.84 & 64.02 & 81.23 & 82.87 & 71.31 & 89.86 & 64.66 & 50.84 & 65.81 & 70.48 \\

       ViTAE-B + VSA(MAE)\cite{wang2022advancing} & Oriented R-CNN\cite{xie2021oriented} & 81.26 & 52.26 & 81.1 & 88.47 & 51.35 & 80.18 & 37.4 & 75.29 & 88.92 & 77.52 & 84.33 & 47.31 & 63.73 & 81.18 & 83.03 & 71.13 & 90.04 & 65.01 & 50.81 & 65.82 & 70.81 \\

       ViT-B + RVSA(MAE)\cite{wang2022advancing} & Oriented R-CNN\cite{xie2021oriented} & 80.92 & 49.88 & 81.05 & 88.52 & 51.52 & 80.17 & 37.87 & 75.96 & 88.83 & 78.46 & 84.01 & 46.53 & 64.18 & 81.21 & 84.04 & 71.34 & 89.99 & 65.41 & 50.53 & 66.49 & 70.85 \\

       ViTAE-B + RVSA(MAE)\cite{wang2022advancing} & Oriented R-CNN\cite{xie2021oriented} & 81.2 & 54.71 & \textbf{\textcolor{blue}{81.12}} & 88.13 & 51.83 & 79.93 & 36.79 & 76.06 & \textbf{\textcolor{blue}{89.23}} & 78.3 & \textbf{\textcolor{blue}{84.46}} & 47.29 & 65.01 & 81.19 & 82.17 & 70.69 & 90.03 & \textbf{\textcolor{red}{66.75}} & 50.73 & 65.4 & 71.05 \\ \hline

       ViT-B12$\times$1(MAE)$\diamondsuit$\cite{wang2022advancing} & FROBB + RT\cite{ding2019learning} & 72.25 & 53.86 & 80.55 & 81.49 & 45.15 & 79.96 & 31.56 & 71.44 & 85.42 & 78.67 & 83.72 & 47.57 & 59.55 & 81.26 & \textbf{\textcolor{blue}{84.87}} & 71.28 & 81.51 & 64.73 & 49.43 & 66.05 & 68.52 \\

       ViT-L12$\times$4(MAE)(Ours) & FROBB + RT\cite{ding2019learning} & 81.2 & \textbf{\textcolor{blue}{60.47}} & 81.01 & \textbf{\textcolor{red}{89.97}} & 51.76 & 80.46 & 39.98 & 78.75 & 89.12 & 78.77 & 84.06 & 53.85 & 60.93 & 81.24 & 84.4 & \textbf{\textcolor{blue}{71.77}} & \textbf{\textcolor{red}{90.15}} & 66.22 & 51.46 & 66.59 & 72.11 \\

       ViT-H12$\times$4(MAE)(Ours) & FROBB + RT\cite{ding2019learning} & \textbf{\textcolor{red}{81.59}} & 55.01 & 80.87 & 88.96 & \textbf{\textcolor{red}{54.27}} & 80.76 & \textbf{\textcolor{red}{40.61}} & \textbf{\textcolor{red}{79.73}} & \textbf{\textcolor{red}{89.47}} & \textbf{\textcolor{red}{79.47}} & 83.84 & \textbf{\textcolor{blue}{55.28}} & \textbf{\textcolor{blue}{65.27}} & \textbf{\textcolor{blue}{89.52}} & 84.42 & \textbf{\textcolor{red}{71.91}} & 90.06 & 65.97 & 51.8 & \textbf{\textcolor{red}{74.12}} & \textbf{\textcolor{blue}{73.15}} \\

       ViT-G12$\times$4(MAE)(Ours) & FROBB + RT\cite{ding2019learning} & \textbf{\textcolor{blue}{81.41}} & \textbf{\textcolor{red}{61.73}} & \textbf{\textcolor{red}{81.17}} & \textbf{\textcolor{blue}{89.86}} & \textbf{\textcolor{blue}{54.12}} & \textbf{\textcolor{blue}{80.84}} & \textbf{\textcolor{blue}{40.35}} & \textbf{\textcolor{blue}{79.42}} & 89.08 & \textbf{\textcolor{blue}{79.3}} & \textbf{\textcolor{red}{84.51}} & \textbf{\textcolor{red}{55.83}} & \textbf{\textcolor{red}{65.61}} & \textbf{\textcolor{red}{89.59}} & \textbf{\textcolor{red}{86.16}} & 71.52 & \textbf{\textcolor{blue}{90.11}} & \textbf{\textcolor{blue}{66.29}} & \textbf{\textcolor{blue}{51.88}} & \textbf{\textcolor{blue}{73.67}} & \textbf{\textcolor{red}{73.62}} \\ \hline

    \end{tabular}
    }
    \label{tab:dior table}
\end{table*}

\autoref{tab:dota2 table} and \autoref{tab:dior table} show the performance results for DOTA v2.0 and DIOR-R, while \autoref{tab:dior sample efficiency table} displays the results of sample efficiency experiments using different amounts of training data. In the tables, red and blue text represent the highest and second-highest performance, respectively. Specifically, the smaller datasets are obtained by randomly sampling 1\%, 5\%, 10\%, 50\%, and 100\% of the images. The number of objects included in each dataset can be seen in \autoref{tab:dior label ratio}.

\textbf{DOTA v2.0}. \autoref{tab:dota2 table} displays the performance results for various models, including our experiments. The results for models other than ours are taken from the official paper \cite{ding2021object}. As noted in the caption of \autoref{tab:dota2 table}, the $\dagger$ represents the result of metrics using the mmrotate framework for fair evaluation. {The abbreviations of method are: Mask R-CNN (MR)\cite{he2017mask}, CMR-Cascade Mask R-CNN (CMR) \cite{chen2019hybrid}, Hybrid Task Cascade without a semanticbranch (HTC) \cite{chen2019hybrid}, Faster R-CNN(FR) \cite{xia2018dota}, Deformable Roi Pooling (Dp) \cite{dai2017deformable} and RoI Transformer (RT) \cite{ding2019learning}. The short names for categories are defined as: PL (Plane), BD (Baseball diamond), BR (Bridge), GTF (Ground track field), SV (Small-vehicle), LV (Large-vehicle), SH (Ship), TC (Tennis-court), BC (Basketball-court), ST (Storage-tank), SBF (Soccer-ball field), RA (Roundabout), HA (Harbor), SP (Swimming-pool), HC (Helicopter), CC (Container-crane), AP (Airport), and HL (Helipad). In backbone column, the IMP means ImageNet classification pretraining. The MAE means MAE on the MillionAID.  Because existing benchmarks have not been performed on mmrotate framework, the $\dagger$ means the result of evaluation using RoI Transformer in Res50 with mmrotate framework. In order to compare the results with the RoI Transformer, $\diamondsuit$ is the result re-implemented in mmrotate framework using the vision transformer weight published by \cite{wang2022advancing}.} Clearly, models pretrained with MAE outperform those pretrained with IMP, with mAP differences ranging from 3.85 to 5.05. In terms of the models proposed in this paper, the relationship between the number of parameters and mAP is such that the mAP tends to increase as the number of parameters increases, with no other changes except for the number of parameters. Although the increase in mAP becomes smaller as the number of model parameters grows, it is evident that performance improves.

\textbf{DIOR-R}. \autoref{tab:dior table} displays the performance results, with the results of models in the table taken from previous research \cite{wang2022advancing}. {The short names for categories are defined as: APL (Airplane), APO (Airport), BF (Baseball field), BC (Baseketball court), BR (Bridge), CH (Chimney), DAM (Dam), ETS (Expressway-toll-station), ESA (Expressway-service-area), GF (Golf field), GTF (Ground track field), HA (Harbor), OP (Overpass), SH(Ship), STA (Stadium), STO (Storage tank), TC (Tennis court), TS (Transition), VE (Vehicle), and WM (Windmill). As same with \autoref{tab:dota2 table}, the IMP means ImageNet classification pretraining. The MAE means MAE on the MillionAID.} Additionally, as shown in \autoref{tab:dior table}, the $\diamondsuit$ represents the result re-implemented in the mmrotate framework using the vision transformer weight published by \cite{wang2022advancing} with the RoI Transformer detection head. As expected, the performance of models pretrained with MAE is higher than those pretrained with IMP. Performance can be improved by changing the backbone structure using the RVSA \cite{wang2022advancing} and ViTAE \cite{xu2021vitae} modules proposed in previous studies. Observing performance changes based solely on the number of parameters in the backbone module, it is clear that mAP increases as the number of backbone parameters grows. Although the rate of improvement in mAP becomes less significant as the number of model parameters increases, it is still evident that performance improves with a higher number of parameters.

\begin{table*}[ht]{\textwidth=0mm}
    \centering
    \caption{the results of class-wise AP and mAP on DIOR-R for evaluating sample efficiency. The short names for categories are defined as same with \autoref{tab:dior table}. In order to compare the results according only backbone, the ViT-B12$\times$1 is retrained by mmrotate framework. The training data of DIOR-R is randomly sampled by ratio 0.01, 0.05, 0.1, 0.5, 1.0. }
    \setlength{\tabcolsep}{2.125pt}
    \renewcommand{\arraystretch}{1.25}
    {\scriptsize
    \begin{tabular}{c | l | c c c c c c c c c c c c c c c c c c c c | c }
        \hline
       Sample ratio & Backbone & APL & APO & BF & BC & BR & CH & DAM & ETS & ESA & GF & GTF & HA & OP & SH & STA & STO & TC & TS & VE & WM & mAP \\ \hline

       \multirow{4}{*}{1$\%$} & ViT-B12$\times$1\cite{wang2022advancing} & 9.10 & \textbf{\textcolor{blue}{0.40}} & 48.20 & 0.00 & 0.00 & 17.70 & 1.30 & 4.50 & 1.40 & 22.10 & 20.40 & 2.10 & 2.80 & 13.20 & \textbf{\textcolor{red}{24.30}} & 29.00 & 55.50 & 8.50 & 10.50 & 5.00 & 13.80 \\
       
        & ViT-L12$\times$4 & \textbf{\textcolor{blue}{10.40}} & \textbf{\textcolor{red}{0.60}} & \textbf{\textcolor{blue}{52.70}} & 0.00 & \textbf{\textcolor{red}{0.70}} & \textbf{\textcolor{blue}{28.00}} & 4.50 & \textbf{\textcolor{blue}{8.20}} & 4.50 & 25.70 & 20.00 & 2.20 & \textbf{\textcolor{blue}{4.50}} & 17.10 & 19.70 & \textbf{\textcolor{red}{42.70}} & 49.00 & \textbf{\textcolor{blue}{10.90}} & \textbf{\textcolor{blue}{11.60}} & \textbf{\textcolor{red}{12.90}} & 16.30 \\
        
        & ViT-H12$\times$4 & 10.20 & 0.30 & 52.20 & \textbf{\textcolor{blue}{0.20}} & 0.00 & \textbf{\textcolor{red}{28.70}} & \textbf{\textcolor{red}{9.10}} & 0.90 & \textbf{\textcolor{red}{9.10}} & \textbf{\textcolor{red}{30.00}} & \textbf{\textcolor{red}{21.80}} & \textbf{\textcolor{blue}{4.50}} & 1.20 & \textbf{\textcolor{blue}{19.00}} & 20.60 & \textbf{\textcolor{blue}{31.10}} & \textbf{\textcolor{blue}{61.60}} & \textbf{\textcolor{blue}{10.90}} & 11.50 & 5.10 & \textbf{\textcolor{blue}{16.40}} \\
        
        & ViT-G12$\times$4 & \textbf{\textcolor{red}{13.70}} & 0.30 & \textbf{\textcolor{red}{53.50}} & \textbf{\textcolor{red}{2.60}} & 0.00 & 20.10 & 3.00 & \textbf{\textcolor{red}{11.20}} & 2.20 & \textbf{\textcolor{blue}{26.00}} & \textbf{\textcolor{blue}{20.80}} & \textbf{\textcolor{red}{6.10}} & \textbf{\textcolor{red}{9.10}} & \textbf{\textcolor{red}{21.10}} & \textbf{\textcolor{blue}{21.40}} & 30.50 
        & \textbf{\textcolor{red}{61.70}} & \textbf{\textcolor{red}{12.10}} & \textbf{\textcolor{red}{12.00}} & \textbf{\textcolor{blue}{12.40}} & \textbf{\textcolor{red}{17.00}} \\ \hline

        \multirow{4}{*}{5$\%$} & ViT-B12$\times$1\cite{wang2022advancing} & \textbf{\textcolor{blue}{61.50}} & \textbf{\textcolor{blue}{6.20}} & \textbf{\textcolor{blue}{63.20}} & \textbf{\textcolor{red}{67.80}} & \textbf{\textcolor{red}{11.70}} & \textbf{\textcolor{blue}{67.20}} & 9.70 & 38.00 & 38.00 & 57.70 & 56.40 & 14.20 & 19.30 & \textbf{\textcolor{red}{62.20}} & 52.10 & \textbf{\textcolor{red}{61.40}} & \textbf{\textcolor{blue}{80.20}} & 20.20 & 33.10 & 33.10 & 42.70 \\
        
        & ViT-L12$\times$4 & 53.10 & 3.60 & 63.00 & \textbf{\textcolor{blue}{66.80}} & 5.80 & \textbf{\textcolor{red}{70.30}} & 4.20 & 35.60 & \textbf{\textcolor{blue}{42.30}} & \textbf{\textcolor{red}{59.80}} & \textbf{\textcolor{blue}{58.30}} & \textbf{\textcolor{red}{18.20}} & 24.10 & 61.90 & \textbf{\textcolor{blue}{55.00}} & \textbf{\textcolor{blue}{61.10}} & 79.30 & \textbf{\textcolor{blue}{21.90}} & \textbf{\textcolor{blue}{33.20}} & \textbf{\textcolor{blue}{41.20}} & 42.90 \\
        
        & ViT-H12$\times$4 & 53.50 & 4.40 & 63.10 & 63.10 & 7.90 & 62.50 & \textbf{\textcolor{blue}{10.40}} & \textbf{\textcolor{red}{39.60}} & \textbf{\textcolor{red}{42.40}} & 58.40 & \textbf{\textcolor{red}{61.50}} & 15.40 & \textbf{\textcolor{blue}{25.50}} & \textbf{\textcolor{blue}{62.10}} & \textbf{\textcolor{red}{58.00}} & 61.00 & \textbf{\textcolor{red}{80.30}} & 19.20 & \textbf{\textcolor{blue}{33.20}} & 40.40 & \textbf{\textcolor{blue}{43.10}} \\
        
        & ViT-G12$\times$4 & \textbf{\textcolor{red}{62.30}} & \textbf{\textcolor{red}{10.20}} & \textbf{\textcolor{red}{63.30}} & 65.60 & \textbf{\textcolor{blue}{11.60}} & 63.40 & \textbf{\textcolor{red}{10.90}} & \textbf{\textcolor{blue}{38.40}} & \textbf{\textcolor{red}{42.40}} & \textbf{\textcolor{blue}{58.60}} & 57.00 & \textbf{\textcolor{blue}{16.80}} & \textbf{\textcolor{red}{27.80}} & 61.30 & 52.00 & 52.60 & 72.40 & \textbf{\textcolor{red}{25.10}} & \textbf{\textcolor{red}{33.40}} & \textbf{\textcolor{red}{41.50}} & \textbf{\textcolor{red}{43.30}} \\ \hline

        \multirow{4}{*}{10$\%$} & ViT-B12$\times$1\cite{wang2022advancing} & 53.50 & 11.70 & 62.90 & 68.60 & 12.90 & 71.70 & 12.50 & 39.70 & 43.10 & 56.80 & 66.10 & 15.20 & 26.70 & 71.30 & 53.20 & 59.60 & 78.80 & 29.30 & 33.10 & 42.90 & 45.50 \\
        
        & ViT-L12$\times$4 & 63.10 & 10.70 & 70.20 & \textbf{\textcolor{red}{79.70}} & 19.80 & \textbf{\textcolor{blue}{71.80}} & 11.50 & 47.60 & \textbf{\textcolor{blue}{52.00}} & 59.20 & 68.20 & 25.50 & \textbf{\textcolor{blue}{38.40}} & \textbf{\textcolor{blue}{79.10}} & \textbf{\textcolor{blue}{63.70}} & 61.40 & \textbf{\textcolor{red}{81.40}} & 32.30 & 39.80 & 51.80 & 51.40 \\
        
        & ViT-H12$\times$4 & \textbf{\textcolor{blue}{71.40}} & \textbf{\textcolor{blue}{15.10}} & \textbf{\textcolor{blue}{71.00}} & \textbf{\textcolor{blue}{79.10}} & \textbf{\textcolor{blue}{22.00}} & 71.60 & \textbf{\textcolor{blue}{15.20}} & \textbf{\textcolor{blue}{49.20}} & 51.10 & \textbf{\textcolor{red}{65.90}} & \textbf{\textcolor{blue}{68.70}} & \textbf{\textcolor{blue}{26.60}} & 36.90 & \textbf{\textcolor{blue}{79.10}} & 62.90 & \textbf{\textcolor{blue}{61.60}} & \textbf{\textcolor{blue}{81.30}} & \textbf{\textcolor{blue}{39.40}} & \textbf{\textcolor{red}{41.10}} & \textbf{\textcolor{blue}{52.50}} & \textbf{\textcolor{blue}{53.10}} \\
        
        & ViT-G12$\times$4 & \textbf{\textcolor{red}{71.90}} & \textbf{\textcolor{red}{15.30}} & \textbf{\textcolor{red}{71.60}} & 77.70 & \textbf{\textcolor{red}{23.70}} & \textbf{\textcolor{red}{72.10}} & \textbf{\textcolor{red}{15.70}} & \textbf{\textcolor{red}{50.80}} & \textbf{\textcolor{red}{52.30}} & \textbf{\textcolor{blue}{64.70}} & \textbf{\textcolor{red}{70.00}} & \textbf{\textcolor{red}{29.30}} & \textbf{\textcolor{red}{42.80}} & \textbf{\textcolor{red}{79.30}} & \textbf{\textcolor{red}{63.90}} & \textbf{\textcolor{red}{61.70}} & 81.00 & \textbf{\textcolor{red}{42.20}} & \textbf{\textcolor{blue}{41.00}} & \textbf{\textcolor{red}{53.00}} & \textbf{\textcolor{red}{54.00}} \\ \hline

        \multirow{4}{*}{50$\%$} & ViT-B12$\times$1\cite{wang2022advancing} & \textbf{\textcolor{blue}{63.40}} & 39.30 & \textbf{\textcolor{blue}{80.20}} & 81.10 & 36.10 & \textbf{\textcolor{red}{72.60}} & 28.00 & 61.40 & 79.00 & 76.40 & 81.70 & 38.20 & 54.40 & \textbf{\textcolor{blue}{81.10}} & 77.90 & 62.90 & \textbf{\textcolor{blue}{81.50}} & 56.70 & 42.90 & 63.50 & 62.90 \\
        
        & ViT-L12$\times$4 & \textbf{\textcolor{red}{72.50}} & 43.20 & 72.40 & 81.00 & 45.00 & \textbf{\textcolor{red}{72.60}} & 29.60 & 70.10 & 86.40 & 75.80 & \textbf{\textcolor{blue}{82.10}} & 45.50 & 58.90 & \textbf{\textcolor{blue}{81.10}} & 77.30 & 63.00 & \textbf{\textcolor{blue}{81.50}} & 63.40 & 50.10 & \textbf{\textcolor{red}{64.90}} & 65.80 \\
        
        & ViT-H12$\times$4 & \textbf{\textcolor{red}{72.50}} & \textbf{\textcolor{blue}{53.40}} & \textbf{\textcolor{red}{80.90}} & \textbf{\textcolor{blue}{81.20}} & \textbf{\textcolor{red}{47.00}} & \textbf{\textcolor{red}{72.60}} & \textbf{\textcolor{red}{36.90}} & \textbf{\textcolor{red}{71.70}} & \textbf{\textcolor{red}{88.70}} & \textbf{\textcolor{red}{77.50}} & \textbf{\textcolor{red}{83.00}} & \textbf{\textcolor{red}{47.80}} & 60.20 & \textbf{\textcolor{blue}{81.10}} & \textbf{\textcolor{red}{83.80}} & \textbf{\textcolor{blue}{71.40}} & \textbf{\textcolor{red}{81.60}} & \textbf{\textcolor{red}{65.40}} & \textbf{\textcolor{red}{51.10}} & \textbf{\textcolor{blue}{64.40}} & \textbf{\textcolor{blue}{68.60}} \\
        
        & ViT-G12$\times$4 & \textbf{\textcolor{red}{72.50}} & \textbf{\textcolor{red}{53.70}} & \textbf{\textcolor{red}{80.90}} & \textbf{\textcolor{red}{81.40}} & \textbf{\textcolor{blue}{46.60}} & \textbf{\textcolor{blue}{72.50}} & \textbf{\textcolor{blue}{35.90}} & \textbf{\textcolor{blue}{71.30}} & \textbf{\textcolor{blue}{88.30}} & \textbf{\textcolor{blue}{76.60}} & \textbf{\textcolor{red}{83.00}} & \textbf{\textcolor{blue}{47.60}} & \textbf{\textcolor{red}{60.30}} & \textbf{\textcolor{red}{89.30}} & \textbf{\textcolor{blue}{83.50}} & \textbf{\textcolor{red}{71.60}} & \textbf{\textcolor{blue}{81.50}} & \textbf{\textcolor{blue}{64.60}} & \textbf{\textcolor{blue}{50.90}} & \textbf{\textcolor{red}{64.90}} & \textbf{\textcolor{red}{68.90}} \\ \hline

        \multirow{4}{*}{100$\%$} & ViT-B12$\times$1\cite{wang2022advancing} & 72.20 & 53.80 & 80.50 & 81.40 & 45.10 & 79.90 & 31.50 & 71.40 & 85.40 & 78.60 & 83.70 & 47.50 & 59.50 & \textbf{\textcolor{blue}{81.20}} & \textbf{\textcolor{blue}{84.80}} & 71.20 & 81.50 & 64.70 & 49.40 & 66.00 & 68.50 \\
        
        & ViT-L12$\times$4 & 81.20 & \textbf{\textcolor{blue}{60.40}} & \textbf{\textcolor{blue}{81.00}} & \textbf{\textcolor{red}{89.90}} & 51.70 & 80.40 & 39.90 & 78.70 & \textbf{\textcolor{blue}{89.10}} & 78.70 & \textbf{\textcolor{blue}{84.00}} & 53.80 & 60.90 & \textbf{\textcolor{blue}{81.20}} & 84.40 & \textbf{\textcolor{blue}{71.70}} & \textbf{\textcolor{red}{90.10}} & \textbf{\textcolor{red}{66.20}} & \textbf{\textcolor{blue}{51.40}} & 66.50 & 72.10 \\
        
        & ViT-H12$\times$4 & \textbf{\textcolor{red}{81.50}} & 55.00 & 80.80 & 88.90 & \textbf{\textcolor{red}{54.20}} & \textbf{\textcolor{blue}{80.70}} & \textbf{\textcolor{red}{40.60}} & \textbf{\textcolor{red}{79.70}} & \textbf{\textcolor{red}{89.40}} & \textbf{\textcolor{red}{79.40}} & 83.80 & \textbf{\textcolor{blue}{55.20}} & \textbf{\textcolor{blue}{65.20}} & \textbf{\textcolor{red}{89.50}} & 84.40 & \textbf{\textcolor{red}{71.90}} & \textbf{\textcolor{blue}{90.00}} & \textbf{\textcolor{blue}{65.90}} & \textbf{\textcolor{red}{51.80}} & \textbf{\textcolor{red}{74.10}} & \textbf{\textcolor{blue}{73.10}} \\

        & ViT-G12$\times$4 & \textbf{\textcolor{blue}{81.40}} & \textbf{\textcolor{red}{61.70}} & \textbf{\textcolor{red}{81.10}} & \textbf{\textcolor{blue}{89.80}} & \textbf{\textcolor{blue}{54.10}} & \textbf{\textcolor{red}{80.80}} & \textbf{\textcolor{blue}{40.30}} & \textbf{\textcolor{blue}{79.40}} & 89.00 & \textbf{\textcolor{blue}{79.30}} & \textbf{\textcolor{red}{84.50}} & \textbf{\textcolor{red}{55.80}} & \textbf{\textcolor{red}{65.60}} & \textbf{\textcolor{red}{89.50}} & \textbf{\textcolor{red}{86.10}} & 71.50 & \textbf{\textcolor{red}{90.10}} & \textbf{\textcolor{red}{66.20}} & \textbf{\textcolor{red}{51.80}} & \textbf{\textcolor{blue}{73.60}} & \textbf{\textcolor{red}{73.60}} \\ \hline

    \end{tabular}
    }
    \label{tab:dior sample efficiency table}
\end{table*}

\begin{table*}[t]{\textwidth=0mm}
    \centering
    \caption{the distribution of instances corresponding to each class for measuring sample efficiency using the DIOR-R dataset. To ensure that the dataset is divided accurately based on images, we need to verify if the objects are distributed according to each ratio. Also, the short names for categories are defined as same with \autoref{tab:dior table}.}
    \setlength{\tabcolsep}{2.75pt}
    \renewcommand{\arraystretch}{1.25}
    {
    \begin{tabular}{c|c c c c c c c c c c c c c c c c c c c c}
        \hline
        \multirow{2}{*}{Sample ratio} & \multicolumn{20}{c}{The Number of Instance} \\ \cline{2-21}
        & APL & APO & BF & BC & BR & CH & DAM & ETS & ESA & GF & GTF & HA & OP & SH & STA & STO & TC & TS & VE & WM \\ \hline

        1 $\%$ & 13 & 8 & 22 & 5 & 17 & 7 & 7 & 7 & 10 & 3 & 14 & 16 & 25 & 260 & 6 & 37 & 50 & 7 & 247 & 19 \\ \hline

        5 $\%$ & 78 & 37 & 101 & 48 & 88 & 34 & 35 & 33 & 51 & 34 & 52 & 105 & 68 & 1195 & 32 & 216 & 220 & 22 & 766 & 136 \\ \hline
        10 $\%$ & 220 & 71 & 234 & 102 & 152 & 72 & 45 & 56 & 90 & 49 & 110 & 319 & 121 & 3040 & 59 & 282 & 601 & 44 & 1323 & 233 \\ \hline
        50 $\%$ & 965 & 325 & 1254 & 560 & 654 & 324 & 221 & 306 & 566 & 258 & 586 & 1244 & 633 & 13296 & 284 & 1730 & 2495 & 253 & 6858 & 1146 \\ \hline
        100 $\%$ & 1888 & 662 & 2384 & 1077 & 1367 & 649 & 512 & 610 & 1080 & 511 & 1162 & 2364 & 1330 & 27351 & 595 & 3042 & 4898 & 501 & 13725 & 2365 \\ \hline
        
    \end{tabular}
    }
    \label{tab:dior label ratio}
\end{table*}

\textbf{Sample Efficiency in DIOR-R}. Sample efficiency in fine-tuning is an important capability that a foundation model should have \cite{brown2020language, zhang2023vitaev2, bommasani2021opportunities, kim2022fine}. In this context, we conduct experiments on the DIOR-R dataset using 1\%, 5\%, 10\%, 50\%, and 100\% of the training data. Each dataset is sampled based on the ratio of images, meaning that they are not sampled precisely based on individual objects. More details about the portion of objects are shown in \autoref{tab:dior label ratio}. To measure sample efficiency, the model changes only the backbone, like in other experiments, and a RoI Transformer is used as a detection head. As shown, the model achieves better performance as the number of parameters increases within the same sample ratio. Additionally, we can verify that using a backbone with a large number of parameters and well-prepared training can achieve comparable performance even if only half of the data is used, as seen with ViT-H12$\times$4, ViT-G12$\times$4 at 50\%, and ViT-B12$\times$1 at 100\%.

These results demonstrate the effectiveness of the proposed models in improving the performance of rotated object detection tasks in remote sensing datasets. The experiments confirm that utilizing larger backbones with more parameters leads to better performance, even though the rate of improvement becomes less significant as the number of parameters increases. Furthermore, the experiments on sample efficiency show that well-prepared training with larger backbones can yield comparable results even when using a smaller portion of the training data, highlighting the importance of optimizing the model architecture and training strategy in order to achieve the best performance possible.

\subsection{Semantic Segmentation}

\begin{table*}[t]{\textwidth=0mm}
    \centering
    \caption{the results of class-wise F1 score, mF1 score and overall accuracy (OA) on Potsdam. As mentioned in dataset detail, the clutter class is not included to evaluate the performance. In order to compare the results with the ViTDET with out any module such as ViTAE and RVSA, $\diamondsuit$ is the result re-implemented in mmsegmentation framework using the vision transformer weight published by \cite{wang2022advancing}.}
    \renewcommand{\arraystretch}{1.0}
    {
    \begin{tabular}{l|c c c c c | c | c}
    
        \hline

       \multirow{2}{*}{Method} & \multicolumn{5}{c|}{F1 score per category ($\%$)} & \multirow{2}{*}{mF1 ($\%$)} & \multirow{2}{*}{OA ($\%$)} \\ \cline{2-6}
        & Imper. surf. & Building & Low veg. & Tree & Car & &  \\ \hline

        FCN \cite{long2015fully} & 88.61 & 93.29 & 83.29 & 79.83 & 93.02 & 87.61 & 85.59 \\
        PSPNet \cite{zhao2017pyramid} & 91.61 & 96.3 & 86.41 & 86.84 & 91.38 & 90.51 & 89.45 \\
        DeeplabV3+ \cite{chen2018encoder} & 92.35 & 96.77 & 85.22 & 86.79 & 93.58 & 90.94 & 89.74 \\
        UperNet \cite{xiao2018unified} & \textbf{\textcolor{red}{93.14}} & 96.75 & 86.3 & 86.13 & 91.56 & 90.78 & 91.26 \\
        S-RA-FCN \cite{mou2020relation} & 91.33 & 94.7 & 86.81 & 83.47 & 94.52 & 90.17 & 88.59 \\ 
        UZ\_1 \cite{volpi2016dense}& 89.3 & 95.4 & 81.8 & 80.5 & 86.5 & 86.7 & 85.8 \\
        UFMG\_4 \cite{nogueira2019dynamic}& 90.8 & 95.6 & 84.4 & 84.3 & 92.4 & 89.5 & 87.9 \\
        Multi-filter CNN \cite{sun2018developing} & 90.94 & \textbf{\textcolor{blue}{96.98}} & 76.32 & 73.37 & 88.55 & 85.23 & 90.65 \\
        HMANet \cite{niu2021hybrid}& 92.38 & 96.08 & \textbf{\textcolor{blue}{86.93}} & 88.21 & 95.44 & 91.81 & 90.46 \\
        LANet \cite{ding2020lanet} & \textbf{\textcolor{blue}{93.05}} & \textbf{\textcolor{red}{97.19}} & \textbf{\textcolor{red}{87.3}} & 88.04 & 94.19 & 91.95 & 90.84 \\
        UperNet-SeCo \cite{manas2021seasonal} & 91.21 & 94.92 & 85.12 & 84.89 & 89.02 & 89.03 & 89.64 \\
        UperNet(RSP-ViTAEv2-S) \cite{wang2022empirical} & \textbf{\textcolor{blue}{93.05}} & 96.62 & 86.62 & 85.89 & 91.01 & 90.64 & 91.21 \\
        UperNet(ViT-B + RVSA) \cite{wang2022advancing} & 92.52 & 96.15 & 86.27 & 85.57 & 90.69 & 90.24 & 90.77 \\
        UperNet(ViTAE-B + RVSA) \cite{wang2022advancing} & 92.97 & 96.7 & 86.68 & 85.92 & 90.93 & 90.64 & 91.22 \\ \hline
        UperNet(ViT-B12$\times$1)$\diamondsuit$\cite{wang2022advancing} & 91.04 & 96.56 & 83.34 & 88.13 & 95.29 & 90.87 & 91.12 \\
        UperNet(ViT-L12$\times$4)(Ours) & 92.22 & 96.98 & 84.75 & \textbf{\textcolor{blue}{88.85}} & 95.92 & 91.75 & 92.17 \\
        UperNet(ViT-H12$\times$4)(Ours) & 92.73 & 96.92 & 85.64 & \textbf{\textcolor{blue}{88.85}} & \textbf{\textcolor{blue}{95.99}} & \textbf{\textcolor{blue}{92.02}} & \textbf{\textcolor{blue}{92.54}} \\
        UperNet(ViT-G12$\times$4)(Ours) & 92.76 & 96.93 & 85.88 & \textbf{\textcolor{red}{89.02}} & \textbf{\textcolor{red}{96.02}} & \textbf{\textcolor{red}{92.12}} & \textbf{\textcolor{red}{92.58}} \\ \hline

    \end{tabular}
    }
    \label{tab:potsdam table}
\end{table*}

\begin{table*}[t]{\textwidth=0mm}
    \centering
    \caption{the results of class-wise IoU and mIoU on LoveDA. In order to compare the results with the ViTDET with out any module such as ViTAE and RVSA, $\diamondsuit$ is the result re-implemented in mmsegmentation framework using the vision transformer weight published by \cite{wang2022advancing}.}
    \renewcommand{\arraystretch}{1.0}
    {
    \begin{tabular}{l|c c c c c c c | c}
    
        \hline

       \multirow{2}{*}{Method} & \multicolumn{7}{c|}{Intersection of Union (IoU)} & \multirow{2}{*}{mIoU} \\ \cline{2-8}
        & Background & Building & Road & Water & Barren & forest & Argriculture &  \\ \hline

        FCN8S \cite{long2015fully} & 42.6 & 49.51 & 48.05 & 73.09 & 11.84 & 43.49 & 58.3 & 46.69 \\
        DeepLabV3+ \cite{chen2018encoder} & 42.97 & 50.88 & 52.02 & 74.36 & 10.4 & 44.21 & 58.53 & 47.62 \\
        PAN \cite{li2018pyramid} & 43.04 & 51.34 & 50.93 & 74.77 & 10.03 & 42.19 & 57.65 & 47.13 \\
        UNet \cite{ronneberger2015u} & 43.06 & 52.74 & 52.78 & 73.08 & 10.33 & 43.05 & 59.87 & 47.84 \\
        UNet++ \cite{zhou2018unet++} & 42.85 & 52.58 & 52.82 & 74.51 & 11.42 & 44.42 & 58.8 & 48.2 \\
        Unetformer \cite{wang2022unetformer} & 44.7 & 58.8 & 54.9 & 79.6 & 20.1 & 46 & 62.5 & 46.9 \\
        Semantic-FPN \cite{kirillov2019panoptic} & 42.93 & 51.53 & 53.43 & 74.67 & 11.21 & 44.62 & 58.68 & 48.15 \\
        PSPNet \cite{zhao2017pyramid} & 44.4 & 52.13 & 53.52 & 76.5 & 9.73 & 44.07 & 57.85 & 48.31 \\
        LinkNet \cite{chaurasia2017linknet} & 43.61 & 52.07 & 52.53 & 76.85 & 12.16 & 45.05 & 57.25 & 48.5 \\
        FarSeg \cite{zheng2020foreground} & 43.09 & 51.48 & 53.85 & 76.61 & 9.78 & 43.33 & 58.9 & 48.15 \\
        FactSeg \cite{ma2021factseg} & 42.6 & 53.63 & 52.79 & 76.94 & 16.2 & 42.92 & 57.5 & 48.94 \\
        HRNet \cite{wang2020deep} & 44.61 & 55.34 & 57.42 & 73.96 & 11.07 & 45.25 & 60.88 & 49.79 \\
        UperNet(ViTAE-B + RVSA) \cite{wang2022advancing} & \textbf{\textcolor{blue}{46.69}} & 58.14 & 57.12 & \textbf{\textcolor{blue}{79.66}} & 16.55 & 46.46 & 62.44 & 52.44 \\ \hline
        UperNet(ViT-B12$\times$1)$\diamondsuit$\cite{wang2022advancing} & 45.69 & 58.75 & 56.7 & 76.56 & 10.56 & \textbf{\textcolor{red}{48.52}} & 62.16 & 51.28 \\
        UperNet(ViT-L12$\times$4) & 46.17 & \textbf{\textcolor{blue}{60.56}} & 57.26 & 76.95 & 16.05 & 47.5 & 62.17 & 52.38 \\
        UperNet(ViT-H12$\times$4) & 46.64 & 59.79 & \textbf{\textcolor{blue}{58.36}} & 79.54 & \textbf{\textcolor{blue}{17.56}} & \textbf{\textcolor{blue}{47.88}} & \textbf{\textcolor{blue}{62.61}} & \textbf{\textcolor{blue}{53.2}} \\
        UperNet(ViT-G12$\times$4) & \textbf{\textcolor{red}{47.57}} & \textbf{\textcolor{red}{61.6}} & \textbf{\textcolor{red}{59.91}} & \textbf{\textcolor{red}{81.79}} & \textbf{\textcolor{red}{18.6}} & 47.3 & \textbf{\textcolor{red}{64}} & \textbf{\textcolor{red}{54.4}} \\ \hline

    \end{tabular}
    }
    \label{tab:loveda table}
\end{table*}

\subsubsection{Dataset}
We evaluate pretrained models for the semantic segmentation downstream task using two benchmark datasets: Potsdam\footnote{\url{https://www.isprs.org/education/benchmarks/UrbanSemLab/2d-sem-label-potsdam.aspx}} and LoveDA.

\textbf{Potsdam}. The Potsdam dataset, published by ISPRS, consists of 38 high-definition images with a pixel resolution of 0.5m. Each image has a fixed pixel size of 6000$\times$6000 and is divided into 24 training images and 14 test images. This dataset is designed for semantic segmentation tasks, classifying six classes at the pixel level: impervious surfaces, buildings, low vegetation, trees, cars, and clutter. The evaluation criteria for the Potsdam dataset are the F1 score and overall accuracy, excluding the clutter category. The dataset is fully published, allowing models to be trained with training images and locally evaluated using test images.

\textbf{LoveDA}. The LoveDA dataset, designed for domain adaptation, features a pixel resolution of 0.3m and consists of 2522 training, 1669 validation, and 1796 test images. The dataset is divided into two areas, rural and urban, and includes seven classes: buildings, roads, water, barren land, forests, agriculture, and background. However, the background class is excluded during learning and evaluation. To demonstrate general performance, we construct the training and test datasets according to the official criteria, disregarding whether images are rural or urban. Unlike Potsdam, performance is estimated using the intersection over union (IoU) across all categories. Since only the test images are published, similar to DOTA v2.0, we evaluate our models by submitting the inference results to the server\footnote{\url{https://codalab.lisn.upsaclay.fr/competitions/421\#learn_the_details-overview}}.

\subsubsection{Implementation Details and Experiment Settings}

\begin{table*}[ht]{\textwidth=0mm}
    \centering
    \caption{the results of F1 score, mF1 score and overall accuracy (OA) for evaluatting sample efficiency in Potsdam. In order to compare the results with the ViTDET with out any module such as ViTAE and RVSA, the ViT-B12$\times$1 is retrained by mmsegmentation framework. The training data of Potsdam is randomly sampled by ratio 0.01, 0.05, 0.1, 0.5, 1.0.}
    \renewcommand{\arraystretch}{1.25}
    {
    \begin{tabular}{c | l | c c c c c | c | c }
        \hline

        \multirow{2}{*}{Sample ratio} & \multirow{2}{*}{Backbone} & \multicolumn{5}{c|}{F1 score per category ($\%$)} & \multirow{2}{*}{mF1 ($\%$)} & \multirow{2}{*}{OA ($\%$)} \\ \cline{3-7}
         &  & Imper. surf. & Building & Low veg. & Tree & Car & & \\ \hline

        \multirow{4}{*}{1$\%$} & ViT-B12$\times$1\cite{wang2022advancing} & 78.27 & 84.54 & 71.04 & 74.66 & 87.23 & 79.15 & 79.43 \\
        & ViT-L12$\times$4 & 79.60 & 84.21 & \textbf{\textcolor{red}{75.13}} & \textbf{\textcolor{red}{79.19}} & 88.44 & \textbf{\textcolor{blue}{81.31}} & \textbf{\textcolor{blue}{81.57}} \\
        & ViT-H12$\times$4 & \textbf{\textcolor{blue}{80.07}} & \textbf{\textcolor{red}{85.43}} & \textbf{\textcolor{red}{75.13}} & 78.19 & \textbf{\textcolor{blue}{88.95}} & \textbf{\textcolor{red}{81.56}} & \textbf{\textcolor{red}{81.78}} \\
        & ViT-G12$\times$4 & \textbf{\textcolor{red}{80.28}} & \textbf{\textcolor{blue}{85.10}} & \textbf{\textcolor{blue}{74.87}} & \textbf{\textcolor{blue}{78.42}} & \textbf{\textcolor{red}{89.13}} & \textbf{\textcolor{red}{81.56}} & \textbf{\textcolor{red}{81.78}} \\ \hline

        \multirow{4}{*}{5$\%$} & ViT-B12$\times$1\cite{wang2022advancing} & 86.15 & \textbf{\textcolor{blue}{92.47}} & 77.14 & 83.49 & 92.35 & 86.32 & 86.34 \\
        & ViT-L12$\times$4 & \textbf{\textcolor{blue}{88.09}} & 92.18 & 81.77 & 84.69 & 93.32 & 88.01 & \textbf{\textcolor{blue}{88.24}} \\
        & ViT-H12$\times$4 & 87.97 & 92.29 & \textbf{\textcolor{blue}{81.86}} & \textbf{\textcolor{blue}{85.05}} & \textbf{\textcolor{blue}{93.75}} & \textbf{\textcolor{blue}{88.19}} & 88.17 \\
        & ViT-G12$\times$4 & \textbf{\textcolor{red}{88.59}} & \textbf{\textcolor{red}{92.85}} & \textbf{\textcolor{red}{82.12}} & \textbf{\textcolor{red}{85.98}} & \textbf{\textcolor{red}{94.24}} & \textbf{\textcolor{red}{88.82}} & \textbf{\textcolor{red}{88.70}} \\ \hline

        \multirow{4}{*}{10$\%$} & ViT-B12$\times$1\cite{wang2022advancing} & 87.91 & 93.77 & 79.72 & 84.99 & 93.93 & 88.06 & 88.32 \\
        & ViT-L12$\times$4 & 89.37 & \textbf{\textcolor{red}{94.49}} & 82.30 & 86.14 & 94.13 & 89.29 & 89.70 \\
        & ViT-H12$\times$4 & \textbf{\textcolor{red}{89.75}} & \textbf{\textcolor{blue}{94.44}} & \textbf{\textcolor{blue}{83.33}} & \textbf{\textcolor{blue}{86.15}} & \textbf{\textcolor{red}{94.89}} & \textbf{\textcolor{blue}{89.71}} & \textbf{\textcolor{red}{90.21}} \\
        & ViT-G12$\times$4 & \textbf{\textcolor{blue}{89.65}} & 94.39 & \textbf{\textcolor{red}{83.80}} & \textbf{\textcolor{red}{86.52}} & \textbf{\textcolor{blue}{94.77}} & \textbf{\textcolor{red}{89.83}} & \textbf{\textcolor{blue}{90.11}} \\ \hline

        \multirow{4}{*}{50$\%$} & ViT-B12$\times$1\cite{wang2022advancing} & 89.65 & 96.20 & 80.96 & 87.26 & 95.15 & 89.84 & 89.95 \\
        & ViT-L12$\times$4 & 92.26 & 96.55 & 85.01 & \textbf{\textcolor{blue}{88.34}} & 96.00 & 91.63 & 92.01 \\
        & ViT-H12$\times$4 & \textbf{\textcolor{red}{92.52}} & \textbf{\textcolor{blue}{96.64}} & \textbf{\textcolor{blue}{85.53}} & 88.19 & \textbf{\textcolor{red}{96.13}} & \textbf{\textcolor{blue}{91.80}} & \textbf{\textcolor{blue}{92.17}} \\
        & ViT-G12$\times$4 & \textbf{\textcolor{blue}{92.35}} & \textbf{\textcolor{red}{96.68}} & \textbf{\textcolor{red}{85.77}} & \textbf{\textcolor{red}{88.56}} & \textbf{\textcolor{blue}{96.05}} & \textbf{\textcolor{red}{91.88}} & \textbf{\textcolor{red}{92.41}} \\ \hline

        \multirow{4}{*}{100$\%$} & ViT-B12$\times$1\cite{wang2022advancing} & 91.04 & 96.56 & 83.34 & 88.13 & 95.29 & 90.87 & 91.22 \\
        & ViT-L12$\times$4 & 92.22 & \textbf{\textcolor{red}{96.98}} & 84.75 & \textbf{\textcolor{blue}{88.85}} & 95.92 & 91.75 & 92.17 \\
        & ViT-H12$\times$4 & \textbf{\textcolor{blue}{92.73}} &\textbf{\textcolor{blue}{ 96.92}} & \textbf{\textcolor{blue}{85.64}} & \textbf{\textcolor{blue}{88.85}} & \textbf{\textcolor{blue}{95.99}} & \textbf{\textcolor{blue}{92.02}} & \textbf{\textcolor{blue}{92.55}} \\
        & ViT-G12$\times$4 & \textbf{\textcolor{red}{92.76}} & 96.93 & \textbf{\textcolor{red}{85.88}} & \textbf{\textcolor{red}{89.02}} & \textbf{\textcolor{red}{96.02}} & \textbf{\textcolor{red}{92.12}} & \textbf{\textcolor{red}{92.59}} \\ \hline

    \end{tabular}
    }
    \label{tab:potsdam sample efficiency table}
\end{table*}

\begin{table}[ht]
    \centering
    \caption{this table shows the distribution of pixel corresponding to each class for measuring sample efficiency using the Potsdam dataset.}
    \setlength{\tabcolsep}{2.75pt}
    \renewcommand{\arraystretch}{1.25}
    {
    \begin{tabular}{c|c c c c c}
        \hline
        \multirow{2}{*}{Sample ratio} & \multicolumn{5}{c}{The Number of Pixel} \\ \cline{2-6}
        & Imper. surf. & Building & Low veg. & Tree & Car \\ \hline

        1 $\%$ & 2480947 & 1945647 & 1903485 & 996186 & 109314 \\ \hline
        5 $\%$ & 11201138 & 10877050 & 8637963 & 5901539 & 475325 \\ \hline
        10 $\%$ & 22595456 & 20900276 & 18146952 & 11961627 & 1009088 \\ \hline
        50 $\%$ & 112858442 & 113779095 & 93552823 & 58452932 & 5139681 \\ \hline
        100 $\%$ & 227566256 & 222620959 & 187276028 & 119657591 & 10574847 \\ \hline
        
    \end{tabular}
    }
    \label{tab:potsdam label ratio}
\end{table}

All experiments benchmarking the proposed backbone are performed using mmsegmentation\footnote{\url{https://github.com/open-mmlab/mmsegmentation}}. As shown in \autoref{subsec:vitdet}, the intermediate features of layers 3, 6, 9, and 12 are resampled to achieve a ratio of 4, 2, 1, 0.5 through each scale block. Both datasets share the same hyperparameters, except for the training iteration. We train the semantic segmentation models using the AdamW optimizer with a base learning rate of 0.00006 and weight decay of 0.01. The learning rate schedule applies the polynomial decay rule with a power of 1.0 and a minimum learning rate of 0. Additionally, linear warm-up is performed during the first 1,500 iterations with a ratio of 0.000001. We apply 160k and 80k training iterations for the Potsdam and LoveDA experiments, respectively. As with rotated object detection, we apply a 0.8 layer-wise learning rate decay and a drop path rate of 0.1 to ViTDET. The UperNet \cite{xiao2018unified}, a popular choice with vision transformer series, is used as the semantic segmentation head.

Since the Potsdam dataset has a fixed pixel size of 6000 $\times$ 6000, we crop the training imagery into patches of size 512 $\times$ 512 with a stride of 384 $\times$ 384. For the LoveDA dataset, with images of size 1024 $\times$ 1024, we configure the data pipeline to randomly crop images into 512 $\times$ 512 patches. For both datasets, we apply data augmentation sequentially, including resize and crop with a ratio range of 0.5 to 2.0, horizontal flip, and photometric distortion. To evaluate full scenes, we perform inference with patches of size 512$\times$512 and a stride of 384 $\times$ 384. For overlapped areas, we set the pixel class based on the average of logits.

\begin{figure*}[p!tb]{}
    \centering
    \includegraphics[width=0.9\textwidth]{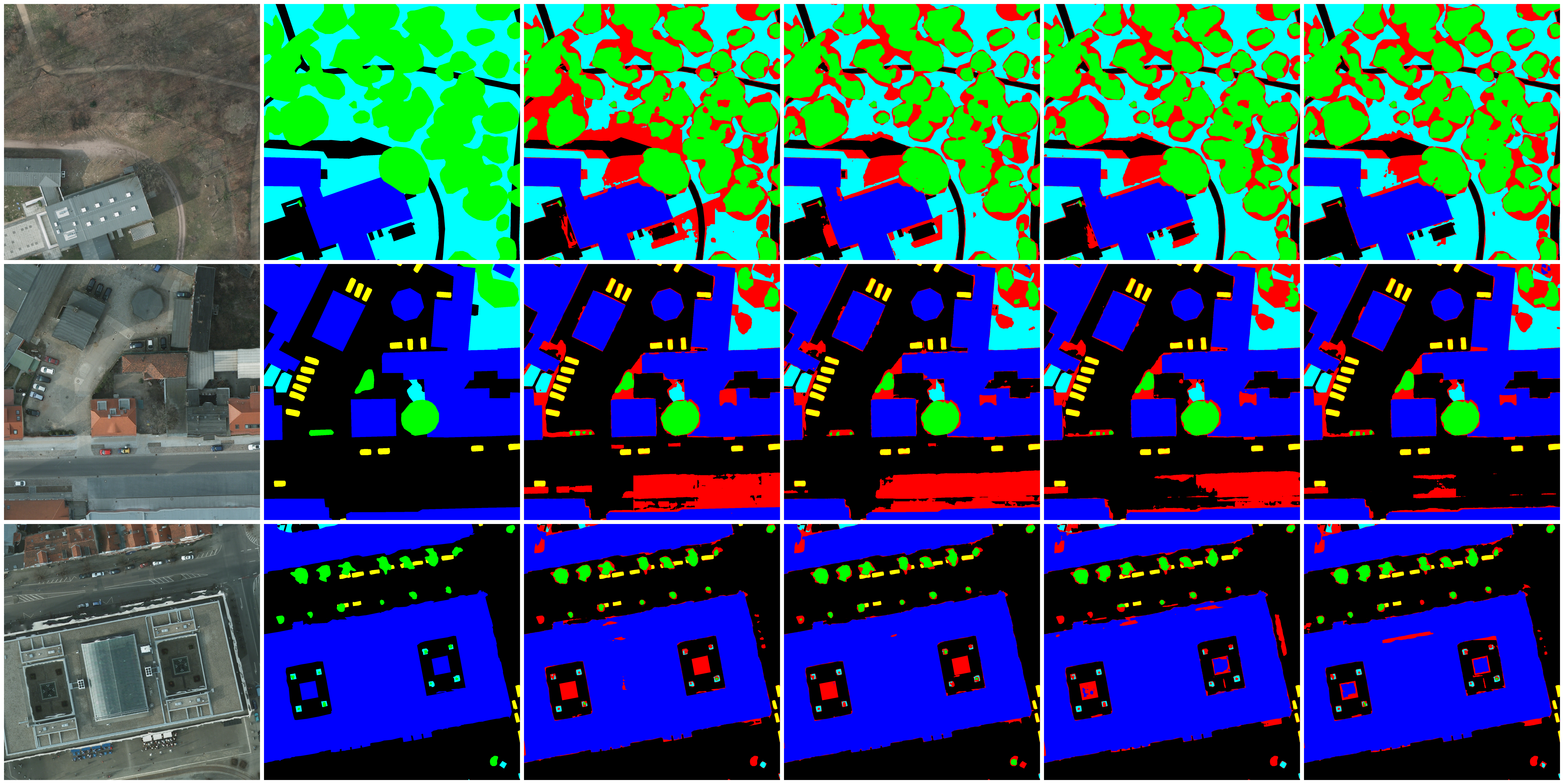}
    \includegraphics[width=0.9\textwidth]{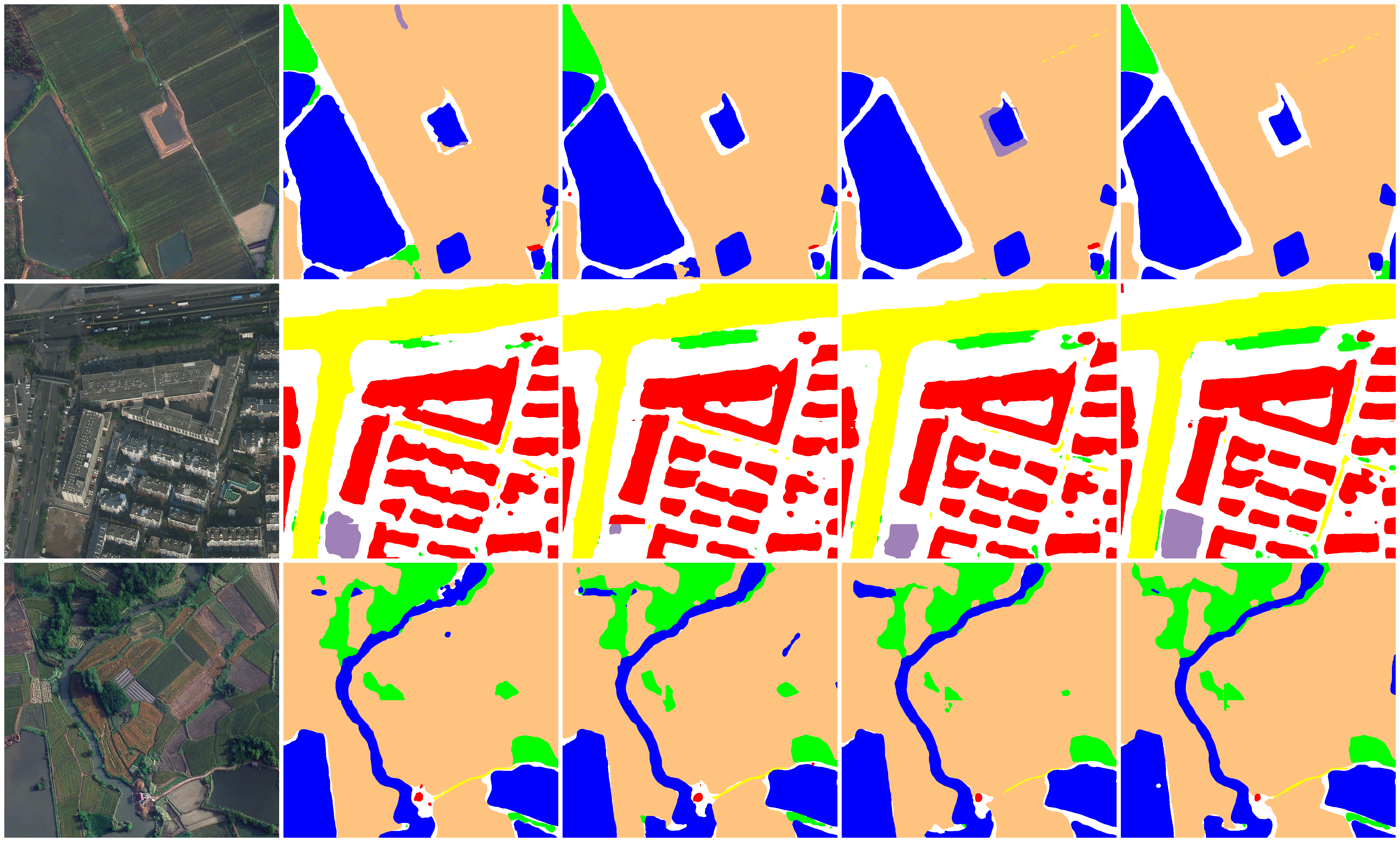}
    \caption{Visualization results of the proposed model. The first to third rows are the results of the Potsdam dataset. The images from left to right are image, label, ViT-B12$\times$1, ViT-L12$\times$4, ViT-H12$\times$4, and ViT-G12$\times$4. The red color in Potsdam dataset means the false prediction. The fourth through sixth rows are the results of the LoveDA. Since the label of test dataset in LoveDA is unavailable, the images from left to right are image, ViT-B12$\times$1, ViT-L12$\times$4, ViT-H12$\times$4, and ViT-G12$\times$4.}
    \label{fig:vis_semantic_seg}
\end{figure*}

\subsubsection{Experiment Results}
Tables \autoref{tab:potsdam table} and \autoref{tab:loveda table} present the semantic segmentation performance results for the Potsdam and LoveDA datasets, respectively. \autoref{tab:potsdam sample efficiency table} demonstrates the improvement in data efficiency as the number of model parameters increases, with fine-tuning results on a reduced number of Potsdam samples. The best performance is indicated in red, while the second-best is in blue. We generate datasets with smaller portions by randomly selecting 1\%, 5\%, 10\%, 50\%, and 100\% of the cropped images from the original dataset. The ratio of pixels representing each category is not accurately allocated, so for precise information on pixel ratios for each category, refer to \autoref{tab:potsdam label ratio}.

\textbf{Potsdam}. \autoref{tab:potsdam table} displays the F1 score, mF1 score, and overall accuracy (OA) for various models using the proposed backbone in Potsdam. As observed in the rotated object detection results, our proposed models outperform those pretrained by IMP in terms of mF1 score and overall accuracy. For the proposed models, a positive correlation exists between the number of parameters and both metrics, indicating that increasing the number of parameters tends to improve performance.

\textbf{LoveDA}. \autoref{tab:loveda table} presents the intersection over union (IoU) results for LoveDA. Although the evaluation metric differs from that of the Potsdam dataset, the performance exhibits a similar pattern.

\textbf{Sample Efficiency in Potsdam}. To evaluate sample efficiency, an essential ability for foundation models in semantic segmentation, we conducted experiments on the Potsdam dataset using varying percentages of training data, including 1\%, 5\%, 10\%, 50\%, and 100\%. As the dataset is partitioned based on images, the pixels corresponding to each class are not divided in an exact ratio. \autoref{tab:potsdam label ratio} details the number of pixels representing each class included in each dataset. The models used for evaluating sample efficiency vary only in the backbone, with UperNet as the segmentation head. As observed, the model performance improves with an increase in the number of parameters while maintaining the same sample ratio. In many cases, using only half or a fifth of the dataset results in superior performance, as seen when comparing the 5\% and 10\% datasets to the 50\% dataset.

\section{Conclusion}

In this study, we have demonstrated that a pretrained model with a vast number of parameters exhibits the essential capabilities expected from a foundation model in the field of modern deep learning. Our findings confirm the efficacy of fine-tuning and sample efficiency in crucial downstream tasks within the remote sensing domain, such as rotated object detection and semantic segmentation.
Moreover, we have established that by employing parallelism in stacking the layers of a vision transformer, the resulting model effectively addresses tasks like rotated object detection and semantic segmentation. Consequently, our results indicate that a remote sensing-oriented foundation model can be achieved by pretraining a vision transformer with a large number of parameters on an extensive dataset of remote sensing images, coupled with the application of parallelism.
Moving forward, our research aims to develop a remote sensing-oriented foundation model using an even more comprehensive collection of remote sensing images, including those from the MillionAID dataset. Furthermore, we plan to investigate strategies for effectively controlling the foundation model through prompt tuning, few-shot learning, and fine-tuning methodologies.

\section*{Acknowledgment}
This work was supported by Artificial intelligence industrial convergence cluster development project funded by the Ministry of Science and ICT (MSIT, Korea) \& Gwangju Metropolitan City. The authors would like to thank Minseok Seo (SI Analytics) and Hakjin Lee (SI Analytics) for initial proofreading and valuable comments.

\ifCLASSOPTIONcaptionsoff
  \newpage
\fi

\bibliographystyle{IEEEtran}
\normalem
\bibliography{vit_rvsa}

\end{document}